\documentclass[journal]{article}
\usepackage{natbib}
\usepackage{amsmath}
\usepackage{amssymb}
\usepackage{amsfonts}
\usepackage{amsthm}
\usepackage{mathrsfs}
\usepackage{latexsym}
\usepackage{multirow}
\usepackage{booktabs}
\usepackage{longtable}
\usepackage{tabularx}
\usepackage{authblk}
\usepackage{caption}
\usepackage{algorithmicx}
\usepackage{algorithm}
\usepackage{algpseudocode}
\usepackage{graphicx}
\usepackage{rotating}
\usepackage{subfigure}
\usepackage{color}
\usepackage{booktabs}
\usepackage{multirow}
\usepackage{adjustbox}
\usepackage{varioref}
\usepackage{tikz}

\allowdisplaybreaks[4]

\usepackage[colorlinks=true,linkcolor=blue]{hyperref}

\newtheorem{theorem}{Theorem}%[section]
\newtheorem{definition}{Definition}%[section]
\newtheorem{remark}{Remark}%[section]

\usepackage{geometry}
\newcommand{\blue}[1]{#1}

\begin{document}
	
\title{\blue{Dual-Core Tensor Wheel Decomposition with Structure-Cached Acceleration  for Tensor Recovery}}
	
\author{Wenchao Xie\thanks{School of Mathematics and Computational Science, Xiangtan University, Xiangtan, 411105, China. Email: youmengx@hotmail.com}, \hskip 0.2cm
Chengcheng Yan \thanks{College of Artificial Intelligence, Shaoxing Institute of Technology, Shaoxing, 312000, China. Email: ycc956176796@gmail.com},  \hskip 0.2cm
Qingsong Wang\thanks{Corresponding author. School of Mathematics and Computational Science, Xiangtan University, National Center for Applied Mathematics in Hunan, Xiangtan, 411105, China. Email: nothing2wang@hotmail.com}, \hskip 0.2cm
% Chengcheng Yan \thanks{School of Mathematics and Computational Science, Xiangtan University, Xiangtan, 411105, China. Email: ycc956176796@gmail.com},  \hskip 0.2cm
% Zheng Peng\thanks{Corresponding author. School of Mathematics and Computational Science, Xiangtan University, Xiangtan, 411105, China. Email: pzheng@xtu.edu.cn} 
}
	
\date{}	
\maketitle
\vspace{-15pt}
\begin{abstract}
Tensor wheel decomposition captures complex correlations in high-dimensional data through interactions between outer-ring factors and a high-order central core. However, updating these components requires the explicit construction and manipulation of large-scale matrices, which limits computational efficiency and incurs particularly high computational and memory costs for the central-core update. To address these limitations, we propose dual-core tensor wheel decomposition (DCTW), which factorizes the central core into two lower-order cores linked by a shared latent bond. Each core is coupled to a different group of outer-ring factors. We further propose a structure-cached acceleration strategy that avoids explicit large-scale matrix multiplications and caches intermediate tensors generated during tensor network contractions for reuse across
block updates. This reuse reduces redundant computation and avoids the explicit construction of large matrices. Complexity analysis and experiments show that the strategy based on structural caching accelerates optimization compared with explicit and implicit update schemes while limiting the increase in peak memory as the ranks grow. Experiments on color video completion and robust denoising show that DCTW matches or outperforms representative tensor network methods in reconstruction quality, computational efficiency, and scalability.
\end{abstract}
	
	
\maketitle
	
\section{Introduction}
As natural higher-order generalizations of matrices, tensors provide a unified representation of multidimensional data, including color videos, remote sensing images, multispectral images, light field images, and traffic data, while preserving complex dependencies across different modes. Tensor decomposition represents or approximates such data through structured interactions among low-dimensional factors, thereby capturing global correlations and reducing the storage and computational costs of high-dimensional data analysis. Over the past several decades, numerous tensor decomposition models have been developed. Classical examples include CANDECOMP/PARAFAC (CP) decomposition \cite{harshman1970foundations,hitchcock1927expression}, Tucker decomposition \cite{tucker1966some}, and tensor singular value decomposition (t-SVD) \cite{kilmer2013third,semerci2014tensor}. More broadly, tensor decomposition methods have been widely applied to signal processing \cite{cichocki2015tensor,zhou2014nonnegative}, machine learning \cite{llosa2023reduced,sidiropoulos2017tensor}, medical imaging \cite{liu2024nonlocal,shen2026unfolding,yokota2019dynamic}, remote sensing image restoration \cite{fang2024cs2dips,mai2024attention}, and light field image compression \cite{ravishankar2022hierarchical,shi2024learning}.

More recently, tensor network (TN) decompositions have emerged as a flexible framework for modeling complex interactions in higher-order data. These methods factorize a high-order tensor into interconnected low-order cores, with the network topology determining how correlations among different modes are modeled. Owing to their ability to characterize multimodal correlations through different topologies, several TN models have been widely applied to tensor recovery in computer vision, particularly tensor completion (TC) \cite{li2025ttucker, zeng2023multimodal} and tensor robust principal component analysis (TRPCA) \cite{chen2025slim, geng2024tensor, wang2023double}.  Among these models, the tensor train (TT) decomposition \cite{oseledets2011tensor} adopts a chain topology, in which successive cores are connected through latent bonds. Specifically, it represents an $N$th-order tensor $\mathcal{X}\in\mathbb{R}^{I_1\times\cdots\times I_N}$ using two boundary matrices $\mathbf{G}_1\in\mathbb{R}^{I_1\times R_1}$ and $\mathbf{G}_N\in\mathbb{R}^{R_{N-1}\times I_N}$, together with third-order cores $\mathcal{G}_k\in\mathbb{R}^{R_{k-1}\times I_k\times R_k}$ for $k=2,\ldots,N-1$. Its element-wise representation is
\begingroup
\setlength{\abovedisplayskip}{4pt}
\setlength{\belowdisplayskip}{4pt}
\setlength{\abovedisplayshortskip}{1pt}
\setlength{\belowdisplayshortskip}{1pt}
$$\mathcal{X}(i_1,\ldots,i_N)
=
\sum_{r_1=1}^{R_1}\cdots\sum_{r_{N-1}=1}^{R_{N-1}}
\mathbf{G}_1(i_1,r_1)
\prod_{k=2}^{N-1}\mathcal{G}_k(r_{k-1},i_k,r_k)
\mathbf{G}_N(r_{N-1},i_N),$$
\endgroup
where $\mathbf{r}_{\mathrm{TT}}=(R_1,\ldots,R_{N-1})$ is the TT-rank. However, the open-chain topology of TT implicitly fixes its boundary ranks to one ($R_0=R_N=1$) and leaves the first and last factors without a direct connection. Tensor ring (TR) decomposition \cite{wang2017efficient,zhao2016tensor} relaxes these boundary constraints by connecting the two ends of the chain, forming a cyclic network of $N$ third-order cores. Specifically, let $\mathcal{G}_k\in\mathbb{R}^{R_{k-1}\times I_k\times R_k}$ for $k=1,\ldots,N$, with $R_0=R_N$. Its element-wise representation is
\begingroup
\setlength{\abovedisplayskip}{4pt}
\setlength{\belowdisplayskip}{4pt}
\setlength{\abovedisplayshortskip}{1pt}
\setlength{\belowdisplayshortskip}{1pt}
$$\mathcal{X}(i_1,\ldots,i_N)
=
\sum_{r_1=1}^{R_1}\cdots\sum_{r_N=1}^{R_N}
\mathcal{G}_1(r_N,i_1,r_1)
\prod_{k=2}^{N}\mathcal{G}_k(r_{k-1},i_k,r_k),$$
\endgroup
and the corresponding TR-rank is $\mathbf{r}_{\mathrm{TR}}=(R_1,\ldots,R_N)$.
Nevertheless, TT and TR connect only adjacent factors, making their representations sensitive to mode ordering and lacking direct interactions between nonadjacent modes. For $N\geq4$, their use of factors of order at most three may also limit the preservation of complex structures in the original high-order tensor subspace. To address these limitations, Zheng et al. proposed the fully-connected tensor network (FCTN) decomposition \cite{zheng2021fully}, which connects every pair of factors through an independent latent bond. Specifically, its $k$th factor is $\mathcal{G}_k\in\mathbb{R}^{R_{1,k}\times\cdots\times R_{k-1,k}\times I_k\times R_{k,k+1}\times\cdots\times R_{k,N}}$ for $k=1,\ldots,N$.  The corresponding element-wise representation is
\begingroup
\setlength{\abovedisplayskip}{4pt}
\setlength{\belowdisplayskip}{4pt}
\setlength{\abovedisplayshortskip}{1pt}
\setlength{\belowdisplayshortskip}{1pt}
$$\mathcal{X}(i_1,\ldots,i_N)
=
\sum_{\{r_{p,q}\}_{1\leq p<q\leq N}}
\prod_{k=1}^{N}
\mathcal{G}_k
\left(r_{1,k},\ldots,r_{k-1,k},i_k,r_{k,k+1},\ldots,r_{k,N}\right),$$
\endgroup
where each $r_{p,q}$ is summed from $1$ to $R_{p,q}$, and index sequences are omitted when empty. The corresponding FCTN-rank is $\mathbf{r}_{\mathrm{FCTN}}=(R_{p,q})_{1\leq p<q\leq N}\in\mathbb{N}^{N(N-1)/2}$.
Although FCTN captures nonlocal correlations and is less sensitive to mode permutations, its $N(N-1)/2$ rank hyperparameters make rank selection costly for higher-order tensors. To capture both local and nonlocal interactions while preserving global structure with fewer rank parameters, Wu et al. proposed tensor wheel (TW) decomposition \cite{wu_tensor_2026}. TW consists of $N$ fourth-order outer-ring factors connected radially to an $N$th-order central core. The ring models local cyclic dependencies, while the core captures nonlocal interactions across modes. Specifically, $\mathcal{G}_k\in\mathbb{R}^{R_k\times I_k\times L_k\times R_{k+1}}$ for $k=1,\ldots,N$, and $\mathcal{C}\in\mathbb{R}^{L_1\times\cdots\times L_N}$, with $R_{N+1}=R_1$. Its element-wise representation is
\begingroup
\setlength{\abovedisplayskip}{4pt}
\setlength{\belowdisplayskip}{4pt}
\setlength{\abovedisplayshortskip}{1pt}
\setlength{\belowdisplayshortskip}{1pt}
$$\mathcal{X}(i_1,\ldots,i_N)
=
\sum_{r_1=1}^{R_1}\cdots\sum_{r_N=1}^{R_N}
\sum_{l_1=1}^{L_1}\cdots\sum_{l_N=1}^{L_N}
\mathcal{C}(l_1,\ldots,l_N)
\prod_{k=1}^{N}\mathcal{G}_k(r_k,i_k,l_k,r_{k+1}),$$
\endgroup
where $r_{N+1}=r_1$, and $R_k$ and $L_k$ are the ring and radial ranks of the $k$th factor, respectively. The TW-rank is $\mathbf{r}_{\mathrm{TW}}=(R_1,\ldots,R_N,L_1,\ldots,L_N)\in\mathbb{N}^{2N}$. Because the number of ring and radial rank parameters grows linearly with the tensor order, TW reduces the cost of rank selection relative to FCTN while retaining both local and nonlocal interactions, providing a practical balance between representation capacity and storage complexity. This advantage, however, does not eliminate the cost of updating the single $N$th-order central core $\mathcal{C}\in\mathbb{R}^{L_1\times\cdots\times L_N}$. Explicitly updating this core typically requires constructing an intermediate tensor of order up to $2N$, whose size depends on both the mode dimensions and the radial ranks. The central-core update therefore becomes a major computational and storage bottleneck, restricting the feasible choices of radial ranks. Ranks that are too small may limit the representation capacity and yield an inadequate decomposition, whereas larger ranks increase computation and memory use.

To address this problem, we propose a dual-core tensor wheel (DCTW) decomposition, as illustrated in Figure~\ref{fig_11}. DCTW replaces the original central core with two cores of lower order connected through a shared latent bond. Each core is coupled to a separate group of factors in the outer ring. The shared bond controls the information exchanged between the two subwheels and can be interpreted as imposing a low-rank factorization on a grouped matricization of the original central core. This design retains the ability of TW to model local and global correlations, introduces a finer hierarchy of ranks, and provides a structure suitable for matrix-free optimization. The main contributions of this paper are summarized as follows.
\begin{itemize}
    \item \textbf{Dual-core tensor wheel topology.}
    \blue{We introduce a dual-core tensor wheel decomposition that replaces the high-order central core of TW with two lower-order cores connected by a shared latent bond and coupled to separate groups of outer-ring factors. DCTW retains the local--global correlation modeling capability of TW, reduces the computation and storage associated with the central core, and provides a flexible hierarchical rank structure for efficient representation.}
   \item \textbf{\blue{Structure-cached acceleration strategy.}}
   We propose a structure-cached acceleration strategy for tensor network decomposition. The strategy avoids explicit large-scale matrix multiplications and reuses intermediate tensors from tensor network contractions across block updates, reducing computational and memory costs while maintaining mathematical equivalence to the explicit-update method. Analyses of floating-point operation counts and memory requirements substantiate its efficiency advantages.
    \item \textbf{\blue{Numerical experiments.}} Extensive experiments on tensor completion using classical and recent color videos and on robust principal component analysis using classical videos show that DCTW achieves high MPSNR and competitive MSSIM. The structure-cached acceleration strategy further enables DCTW to outperform the other evaluated tensor network decomposition methods in runtime.

\end{itemize}

The remainder of this paper is organized as follows. Section~\ref{sect_2} introduces the notation, definitions, and relevant properties. Section~\ref{sect_3} presents the proposed model and solution algorithm. Section~\ref{sect_4} describes the implicit-update and structure-cached acceleration strategies and analyzes their computational and storage complexities. Section~\ref{sect_5} evaluates the iterative speedup and memory use of structure caching and reports experimental results on tensor completion and tensor robust principal component analysis. Section~\ref{sect_6} concludes the paper.

\vspace{-2pt}
% %%%%%%%%%%%%%%%%%%%%
\section{Notations and Preliminaries}\label{sect_2}

\subsection{Notations}
In this paper, we primarily follow the notation conventions of \cite{kolda2009tensor}, using bold lowercase letters (e.g., $\mathbf{x}$), bold uppercase letters (e.g., $\mathbf{X}$), and calligraphic letters (e.g., $\mathcal{X}$) to denote vectors, matrices, and tensors, respectively. Given a matrix $\mathbf{X}\in\mathbb{R}^{I\times I}$, we define its trace as $\operatorname{tr}(\mathbf{X})=\sum_{i=1}^{I}x_{ii}$, where $x_{ii}$ denotes the $i$th diagonal element of $\mathbf{X}$. For an $N$th-order tensor $\mathcal{X}\in\mathbb{R}^{I_1\times\cdots\times I_N}$, we use $\mathcal{X}(i_1,i_2,\ldots,i_N)$ to denote its $(i_1,i_2,\ldots,i_N)$th element and define its Frobenius norm and $\ell_1$-norm as $\lVert\mathcal{X}\rVert_F:=\left(\sum_{i_1,\ldots,i_N}\left|\mathcal{X}(i_1,\ldots,i_N)\right|^2\right)^{1/2}$ and $\lVert\mathcal{X}\rVert_1:=\sum_{i_1,\ldots,i_N}\left|\mathcal{X}(i_1,\ldots,i_N)\right|$, respectively. The mode-$k$ unfolding of $\mathcal{X}$ and the corresponding folding operation are denoted by $\mathbf{X}_{(k)}=\operatorname{Unfold}_k(\mathcal{X})$ and $\mathcal{X}=\operatorname{Fold}_k(\mathbf{X}_{(k)})$, respectively.

\subsection{Preliminaries}
\begin{definition}[Generalized Tensor Transposition {\cite{zheng2021fully}}]
Let $\mathcal{X}\in
\mathbb{R}^{I_1\times I_2\times\cdots\times I_N}$ be an $N$th-order tensor, and let
$\mathbf{n}=(n_1,n_2,\ldots,n_N)$ be an arbitrary permutation of
$(1,2,\ldots,N)$. The tensor obtained by permuting the modes of
$\mathcal{X}$ according to $\mathbf{n}$ is called the generalized tensor
transposition, or the $\mathbf{n}$-mode transposition, of $\mathcal{X}$. It is denoted by $\vec{\mathcal{X}}^{\,\mathbf{n}}
\in
\mathbb{R}^{
I_{n_1}\times I_{n_2}\times\cdots\times I_{n_N}
}.$ Its entries satisfy $\vec{\mathcal{X}}^{\,\mathbf{n}}
\left(i_{n_1},i_{n_2},\ldots,i_{n_N}\right)
=
\mathcal{X}(i_1,i_2,\ldots,i_N)$ for all $i_k\in\{1,2,\ldots,I_k\}$ and $k=1,2,\ldots,N$.
\end{definition}

\begin{definition}[Generalized Tensor Unfolding {\cite{zheng2021fully}}]
\label{def:generalized_tensor_unfolding}

Given an $N$th-order tensor $\mathcal{X}
\in
\mathbb{R}^{I_1\times I_2\times\cdots\times I_N}$, let $\mathbf{n}=(n_1,n_2,\ldots,n_N)$ be an arbitrary permutation of $(1,2,\ldots,N)$. The generalized
tensor unfolding of $\mathcal{X}$ is defined as the matrix
\[
\mathbf{X}_{[n_{1:d}\mid n_{d+1:N}]}
\in
\mathbb{R}^{
\left(\prod_{i=1}^{d}I_{n_i}\right)
\times
\left(\prod_{i=d+1}^{N}I_{n_i}\right)
},
\]
whose elements satisfy
\[
\mathbf{X}_{[n_{1:d}\mid n_{d+1:N}]}(j_1,j_2)
=
\mathcal{X}(i_1,i_2,\ldots,i_N),
\]
where
\[
j_1
=
i_{n_1}
+
\sum_{s=2}^{d}
\left[
\left(i_{n_s}-1\right)
\prod_{m=1}^{s-1}I_{n_m}
\right]\quad
and \quad
j_2
=
i_{n_{d+1}}
+
\sum_{s=d+2}^{N}
\left[
\left(i_{n_s}-1\right)
\prod_{m=d+1}^{s-1}I_{n_m}
\right].
\]
The generalized unfolding operation and its inverse are denoted,
respectively, by
$\mathbf{X}_{[n_{1:d}\mid n_{d+1:N}]}
=
\operatorname{GUnfold}
\left(
\mathcal{X},
n_{1:d}\mid n_{d+1:N}
\right)$
and $\mathcal{X}
=
\operatorname{GFold}
\left(
\mathbf{X}_{[n_{1:d}\mid n_{d+1:N}]},
n_{1:d}\mid n_{d+1:N}
\right).$

\end{definition}

Two useful special cases follow directly from Definition~\ref{def:generalized_tensor_unfolding} by setting $d=1$ and choosing appropriate mode permutations. For $\mathbf{n}=(k,1{:}k-1,k+1{:}N)$, the generalized unfolding reduces to the classical mode-$k$ unfolding, denoted by

$$
\mathbf{X}_{(k)}
:=
\mathbf{X}_{[k\mid 1{:}k-1,\,k+1{:}N]}.
$$
For the cyclic permutation $\mathbf{n}=(k{:}N,1{:}k-1)$, it reduces to the cyclic mode-$k$ unfolding used in TR decomposition, denoted by
$$
\mathbf{X}_{\langle k\rangle}
:=
\mathbf{X}_{[k\mid k+1{:}N,\,1{:}k-1]}.
$$
The inverse operators of these two unfolding operations are denoted by $\operatorname{Fold}_k(\cdot)$ and $\operatorname{Fold}_{\langle k\rangle}(\cdot)$, respectively.

% \begin{definition}[Tensor Contraction {\cite{zheng2021fully}}]
% \label{def3}
% Let $\mathcal{X}\in\mathbb{R}^{I_1\times I_2\times\cdots\times I_N}$
% , $\mathcal{Y}\in\mathbb{R}^{J_1\times J_2\times\cdots\times J_M}$ and suppose that $I_{n_i}=J_{m_i}$ for $i=1,2,\ldots,d.$ Here,
% $\mathbf{n}=(n_1,n_2,\ldots,n_N)$ and
% $\mathbf{m}=(m_1,m_2,\ldots,m_M)$ are permutations of 
% $(1,2,\ldots,N)$ and $(1,2,\ldots,M)$, respectively, satisfying $n_{d+1}<n_{d+2}<\cdots<n_N$ and $m_{d+1}<m_{d+2}<\cdots<m_M$.
% The contraction between modes
% $n_{1:d}=(n_1,\ldots,n_d)$ of $\mathcal{X}$ and modes
% $m_{1:d}=(m_1,\ldots,m_d)$ of $\mathcal{Y}$ is denoted by
% \[
% \mathcal{Z}
% =
% \mathcal{X}
% \times_{n_{1:d}}^{m_{1:d}}
% \mathcal{Y}.
% \]
% The resulting tensor is of order $N+M-2d$ and satisfies
% $\mathcal{Z}
% \in
% \mathbb{R}^{
% I_{n_{d+1}}\times\cdots\times I_{n_N}
% \times
% J_{m_{d+1}}\times\cdots\times J_{m_M}
% }$. Equivalently, using the generalized tensor unfolding notation, the
% contraction can be expressed as
% $$\mathbf{Z}_{
% [\,1:N-d\,;\,N-d+1:N+M-2d\,]
% }
% =
% \mathbf{X}_{
% [\,n_{d+1:N}\,;\,n_{1:d}\,]
% }
% \mathbf{Y}_{
% [\,m_{1:d}\,;\,m_{d+1:M}\,]
% }.$$
% \end{definition}

\begin{definition}[Tensor Contraction {\cite{zheng2021fully}}]
\label{def3}
Let $\mathcal{X}\in\mathbb{R}^{I_1\times I_2\times\cdots\times I_N}$,
$\mathcal{Y}\in\mathbb{R}^{J_1\times J_2\times\cdots\times J_M}$
and suppose that $I_{n_i}=J_{m_i}$ for $i=1,2,\ldots,d$.
Here,
$\mathbf{n}=(n_1,n_2,\ldots,n_N)$ and
$\mathbf{m}=(m_1,m_2,\ldots,m_M)$ are permutations of
$(1,2,\ldots,N)$ and $(1,2,\ldots,M)$, respectively, satisfying
$n_{d+1}<n_{d+2}<\cdots<n_N$ and
$m_{d+1}<m_{d+2}<\cdots<m_M$. The contraction between modes
$n_{1:d}=(n_1,\ldots,n_d)$ of $\mathcal{X}$ and modes
$m_{1:d}=(m_1,\ldots,m_d)$ of $\mathcal{Y}$ is denoted by
\[
\mathcal{Z}
=
\mathcal{X}
\times_{n_{1:d}}^{m_{1:d}}
\mathcal{Y}.
\]
The resulting tensor is of order $N+M-2d$ and satisfies
$\mathcal{Z}
\in
\mathbb{R}^{
I_{n_{d+1}}\times\cdots\times I_{n_N}
\times
J_{m_{d+1}}\times\cdots\times J_{m_M}
}.$ Equivalently, using the generalized tensor unfolding notation, the
contraction can be expressed as
\[
\mathbf{Z}_{[1:N-d\,;\,N-d+1:N+M-2d]}
=
\mathbf{X}_{[n_{d+1:N}\,;\,n_{1:d}]}
\mathbf{Y}_{[m_{1:d}\,;\,m_{d+1:M}]}.
\]
\end{definition}

\begin{figure}[!ht]
    \centering
    \includegraphics[width=0.85\linewidth]{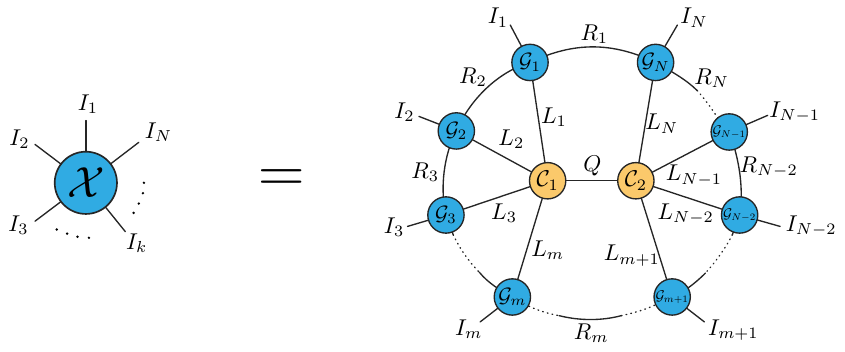}
    \caption{The Dual-Core Tensor Wheel.}
    \label{fig_11}
\end{figure}

\begin{definition}[Dual-Core Tensor Wheel]
\label{def:dctw}
Let $\mathcal{X}\in\mathbb{R}^{I_1\times\cdots\times I_N}$ be an $N$th-order tensor with $N\geq 3$, and let $m=\lceil N/2\rceil$. The modes are partitioned into two ordered, consecutive groups
$$
\mathbf{s}_1=(1,\ldots,m),\qquad
\mathbf{s}_2=(m+1,\ldots,N),
$$
where $|\mathbf{s}_1|=\lceil N/2\rceil$ and $|\mathbf{s}_2|=\lfloor N/2\rfloor$. As illustrated in Figure~\ref{fig_11}, the DCTW decomposition represents $\mathcal{X}$ using $N$ fourth-order outer-ring factors
$$
\mathcal{G}_k
\in
\mathbb{R}^{R_k\times I_k\times L_k\times R_{k+1}},
\qquad k=1,\ldots,N,
$$
where $R_{N+1}=R_1$, and two central cores
$$
\mathcal{C}_1
\in
\mathbb{R}^{L_1\times\cdots\times L_m\times Q},
\qquad
\mathcal{C}_2
\in
\mathbb{R}^{Q\times L_{m+1}\times\cdots\times L_N}.
$$
Here, $R_k$ is the $k$th ring rank, $L_k$ is the $k$th radial rank, and $Q$ is the shared rank connecting the two central cores. The element-wise representation of the DCTW decomposition is
$$
\begin{aligned}
\mathcal{X}(i_1,\ldots,i_N)
={}&
\sum_{r_1=1}^{R_1}\cdots
\sum_{r_N=1}^{R_N}
\sum_{q=1}^{Q}
\left[
\sum_{l_1=1}^{L_1}\cdots
\sum_{l_m=1}^{L_m}
\mathcal{C}_1(l_1,\ldots,l_m,q)
\prod_{k=1}^{m}
\mathcal{G}_k(r_k,i_k,l_k,r_{k+1})
\right]
\\
&\times
\left[
\sum_{l_{m+1}=1}^{L_{m+1}}\cdots
\sum_{l_N=1}^{L_N}
\mathcal{C}_2(q,l_{m+1},\ldots,l_N)
\prod_{k=m+1}^{N}
\mathcal{G}_k(r_k,i_k,l_k,r_{k+1})
\right],
\qquad r_{N+1}=r_1.
\end{aligned}
\label{eq:dctw_element}
$$
The DCTW rank is defined as
$$
\mathbf{r}_{\mathrm{DCTW}}
=
(\mathbf{R},\mathbf{L},Q),
\qquad
\mathbf{R}=(R_1,\ldots,R_N),
\qquad
\mathbf{L}=(L_1,\ldots,L_N),
$$
and therefore contains $2N+1$ rank parameters. The tensor contraction form of DCTW is
$$
\begin{aligned}
\mathcal{X}
={}&
\Bigg(
\mathcal{G}_{1}
\times_{4}^{1}\mathcal{G}_{2}
\times_{6}^{1}\cdots
\times_{2m}^{1}\mathcal{G}_{m}
\times_{(3,5,\ldots,2m+1)}^{(1,2,\ldots,m)}
\mathcal{C}_1
\Bigg)
\\
&\times_{(1,m+2,m+3)}^{(N-m+3,2,1)}
\Bigg(
\mathcal{C}_2
\times_{(2,3,\ldots,N-m+1)}^{(3,5,\ldots,2(N-m)+1)}
\Bigg(
\mathcal{G}_{{m+1}}
\times_{4}^{1}\mathcal{G}_{m+2}
\times_{6}^{1}\cdots
\times_{2(N-m)}^{1}\mathcal{G}_{N}
\Bigg)
\Bigg).
\end{aligned}
\label{eq:dctw_tensor_form}
$$
We denote the DCTW decomposition by
$$
\operatorname{DCTW}
\left(
\{\mathcal{G}_k\}_{k=1}^{m};
\mathcal{C}_1
\overset{Q}{\longleftrightarrow}
\mathcal{C}_2;
\{\mathcal{G}_k\}_{k=m+1}^{N}
\right),
$$
where $\mathcal{C}_1$ and $\mathcal{C}_2$ are connected by a shared latent bond of dimension $Q$. More compactly, it is written as
$$
\operatorname{DCTW}
\left(
\{\mathcal{G}_{\pi_k}\}_{k=1}^{N};
\mathcal{C}_1,\mathcal{C}_2
\right).
$$
\end{definition}

\begin{remark}
\label{rem:shared_rank_position}
The shared mode of dimension $Q$ may occupy any position in either central core. For example,
$$
\begin{aligned}
\mathcal{C}_1
&\in
\mathbb{R}^{L_1\times\cdots\times L_a\times Q\times L_{a+1}\times\cdots\times L_m},
&&1\leq a<m,\\
\mathcal{C}_2
&\in
\mathbb{R}^{L_{m+1}\times\cdots\times L_b\times Q\times L_{b+1}\times\cdots\times L_N},
&&m+1\leq b<N.
\end{aligned}
$$
These forms differ from Definition~\ref{def:dctw} only by permutations of the central-core modes and yield the same DCTW representation. For convenience, we place $Q$ last in $\mathcal{C}_1$ and first in $\mathcal{C}_2$, denoted by $\mathcal{C}_1\overset{Q}{\longleftrightarrow}\mathcal{C}_2$, to keep the element-wise and tensor contraction forms compact.
\end{remark}

\begin{definition}[Dimension-Set Tensor Contraction]
\label{def:dimension_set_contraction}
Let
$\mathcal{G}_k\in\mathbb{R}^{R_k\times I_k\times L_k\times R_{k+1}}$,
$k=1,\ldots,N$, be the fourth-order outer-ring factors, where
$R_{N+1}=R_1$. Let
$\mathcal{C}_1\in\mathbb{R}^{L_1\times\cdots\times L_m\times Q}$
and
$\mathcal{C}_2\in\mathbb{R}^{Q\times L_{m+1}\times\cdots\times L_N}$
be the two central cores. Define
$$
\Omega
=
\left\{
(\alpha_i,\widehat{\alpha}_i)
\mid i=1,\ldots,d
\right\}
$$
as a set of compatible mode pairs. The left half-ring tensor is obtained by the dimension-set contraction
$$
\mathcal{M}_{\mathrm L}
=
\operatorname{contract}
\left(
\mathcal{G}_1,\ldots,\mathcal{G}_m,\mathcal{C}_1;
\Omega_{\mathrm L}
\right),
$$
where $\mathcal{M}_{\mathrm L}\in\mathbb{R}^{R_1\times I_1\cdots\times I_m\times R_{m+1}\times Q}$ and 
$$
\begin{aligned}
\Omega_{\mathrm L}
={}&
\left\{
\left(
R_{k+1}^{(\mathcal{G}_k)},
R_{k+1}^{(\mathcal{G}_{k+1})}
\right)
\mid k=1,\ldots,m-1
\right\}
\cup
\left\{
\left(
L_k^{(\mathcal{G}_k)},
L_k^{(\mathcal{C}_1)}
\right)
\mid k=1,\ldots,m
\right\}.
\end{aligned}
$$
Equivalently,
$$
\mathcal{M}_{\mathrm L}
=
\left(
\left(
\left(
\mathcal{G}_1
\times_{4}^{1}
\mathcal{G}_2
\right)
\times_{6}^{1}
\mathcal{G}_3
\right)
\cdots
\times_{2m}^{1}
\mathcal{G}_m
\right)
\times_{(3,5,\ldots,2m+1)}^{(1,2,\ldots,m)}
\mathcal{C}_1.
$$

Suppose that an intermediate factor $\mathcal{G}_j$,
$2\leq j\leq m-1$, is excluded from the left half ring. The remaining tensors form the environment tensor
$$
\mathcal{M}_{\mathrm L_{\neq j}}
=
\operatorname{contract}
\left(
\mathcal{G}_1,\ldots,\mathcal{G}_{j-1},
\mathcal{G}_{j+1},\ldots,\mathcal{G}_m,
\mathcal{C}_1;
\Omega_{\mathrm L_{\neq j}}
\right),
$$
where $\mathcal{M}_{\mathrm L_{\neq j}}
\in
\mathbb{R}^{
R_1
\times I_1\times\cdots\times I_{j-1}
\times R_j
\times L_j
\times Q
\times R_{j+1}
\times I_{j+1}\times\cdots\times I_m
\times R_{m+1}
}$ and
$$
\begin{aligned}
\Omega_{\mathrm L_{\neq j}}
={}&
\left\{
\left(
R_{k+1}^{(\mathcal{G}_k)},
R_{k+1}^{(\mathcal{G}_{k+1})}
\right)
\;\middle|\;
\begin{array}{c}
k=1,\ldots,m-1,\\
k\notin\{j-1,j\}
\end{array}
\right\}
\cup
\left\{
\left(
L_k^{(\mathcal{G}_k)},
L_k^{(\mathcal{C}_1)}
\right)
\;\middle|\;
\begin{array}{c}
k=1,\ldots,m,\\
k\neq j
\end{array}
\right\}.
\end{aligned}
$$
An equivalent sequential contraction is
$$
\begin{aligned}
\mathcal{M}_{\mathrm L_{\neq j}}
={}&
\left[
\Bigl(\bigl((\mathcal{G}_1\times_4^1\mathcal{G}_2)\times_6^1\mathcal{G}_3\cdots\bigr)\times_{2j-2}^1\mathcal{G}_{j-1}\Bigr)
\times_{(3,5,\ldots,2j-1)}^{(1,2,\ldots,j-1)}
\mathcal{C}_1
\right]\\
&\times_{(j+3,j+4,\ldots,m+2)}^{(3,5,\ldots,2(m-j)+1)}
\Bigl(\bigl((\mathcal{G}_{j+1}\times_4^1\mathcal{G}_{j+2})\times_6^1\mathcal{G}_{j+3}\cdots\bigr)\times_{2(m-j)}^1\mathcal{G}_m\Bigr).
\end{aligned}
$$
For $j=2$ or $j=m-1$, the corresponding single-factor chain is interpreted as $\mathcal{G}_1$ or $\mathcal{G}_m$, respectively. Only the left half-ring contraction is presented here. The right half-ring contraction follows analogously by replacing $\mathcal{C}_1$, $\{\mathcal{G}_k\}_{k=1}^{m}$, and $\Omega_{\mathrm L}$ with $\mathcal{C}_2$, $\{\mathcal{G}_k\}_{k=m+1}^{N}$, and $\Omega_{\mathrm R}$, respectively.
\end{definition}

\begin{remark}
Dimension-set tensor contraction is an explicit notation for generalized tensor contraction that identifies the contracted mode pairs without introducing a new tensor operation. Computationally, the tensors are permuted and unfolded into compatible matrices, multiplied, and then folded back into a tensor.
\end{remark}

Both generalized tensor contraction and dimension-set tensor contraction are used throughout the following sections. When only a few modes are contracted, we list them directly and omit explicit mode pairings for simplicity.

\begin{theorem}[DCTW Subwheel Equations]
\label{the1}
Assume that $\mathcal{X}\in\mathbb{R}^{I_1\times\cdots\times I_N}$ admits the generalized DCTW decomposition
$$
\mathcal{X}
=
\operatorname{DCTW}
\left(
\{\mathcal{G}_{\pi_k}\}_{k=1}^{N};
\mathcal{C}_1,\mathcal{C}_2
\right).
\label{eq:dctw_decomposition}
$$
Let $\mathbf{s}_1=(1,\ldots,m)$ and $\mathbf{s}_2=(m+1,\ldots,N)$. For the $k$th outer-ring factor $\mathcal{G}_k$ associated with $\mathbf{s}_1$, where $1\leq k\leq m$, the subwheel equation is
$$
\mathbf{X}_{\langle k\rangle}
=
\operatorname{GUnfold}
\left(
\mathcal{G}_k,
2\mid1,3,4
\right)
\operatorname{GUnfold}
\left(
\mathcal{M}_{\neq k},
k{:}k+2
\mid
k+3{:}N+2,1{:}k-1
\right).
\label{eq:tensor_subwheel}
$$
Here,
$
\mathcal{M}_{\neq k}
\in
\mathbb{R}^{
I_1\times\cdots\times I_{k-1}
\times R_k\times L_k\times R_{k+1}
\times I_{k+1}\times\cdots\times I_N
}
$
is the $(N+2)$th-order subwheel tensor obtained by contracting all DCTW factors except $\mathcal{G}_k$. Its tensor-contraction form is
$$
\begin{aligned}
\mathcal{M}_{\neq k}
={}&
\Bigg[
\Bigg(
\mathcal{G}_1
\times_{4}^{1}\mathcal{G}_2
\times_{6}^{1}\cdots
\times_{2(k-1)}^{1}\mathcal{G}_{k-1}
\Bigg)
\times_{(3,5,\ldots,2k-1)}^{(1,2,\ldots,k-1)}
\mathcal{C}_1
\\[-1mm]
&\qquad
\times_{(k+3,k+4,\ldots,m+2)}^{(3,5,\ldots,2(m-k)+1)}
\Bigg(
\mathcal{G}_{k+1}
\times_{4}^{1}\mathcal{G}_{k+2}
\times_{6}^{1}\cdots
\times_{2(m-k)}^{1}\mathcal{G}_m
\Bigg)
\Bigg]
\\[-1mm]
&\times_{(1,m+5,k+3)}^{(N-m+3,2,1)}
\Bigg[
\mathcal{C}_2
\times_{(2,3,\ldots,N-m+1)}^{(3,5,\ldots,2(N-m)+1)}
\Bigg(
\mathcal{G}_{m+1}
\times_{4}^{1}\mathcal{G}_{m+2}
\times_{6}^{1}\cdots
\times_{2(N-m)}^{1}\mathcal{G}_N
\Bigg)
\Bigg].
\end{aligned}
\label{eq:dctw_subwheel_tensor_form}
$$
A contraction chain containing a single factor reduces to that factor. Empty chains and their associated mode lists are omitted. The same construction applies to an outer-ring factor associated with $\mathbf{s}_2$, where $m+1\leq k\leq N$.

For the first central core $\mathcal{C}_1$, the subwheel equation is
$$
\mathbf{x}
=
\mathbf{c}_1
\mathbf{N}_{\neq\mathcal{C}_1}.
\label{eq:c1_subwheel}
$$
The matrices and vectors are defined as
$$
\begin{aligned}
\mathbf{N}_{\neq\mathcal{C}_1}
&=
\operatorname{GUnfold}
\left(
\mathcal{N}_{\neq\mathcal{C}_1},
2,4,\ldots,2m,2m+1
\mid
1,3,\ldots,2m-1,2m+2{:}N+m+1
\right),\\
\mathbf{x}
&=
\operatorname{GUnfold}
\left(
\mathcal{X},
\varnothing\mid1{:}N
\right),
\mathbf{c}_1
=
\operatorname{GUnfold}
\left(
\mathcal{C}_1,
\varnothing\mid1{:}m+1
\right).
\end{aligned}
$$
The environment tensor
$
\mathcal{N}_{\neq\mathcal{C}_1}
\in
\mathbb{R}^{
I_1\times L_1\times\cdots\times
I_m\times L_m\times Q\times
I_{m+1}\times\cdots\times I_N
}
$
is an $(N+m+1)$th-order tensor with contraction form
$$
\begin{aligned}
\mathcal{N}_{\neq\mathcal{C}_1}
={}&
\Bigg(
\mathcal{G}_1
\times_{4}^{1}\mathcal{G}_2
\times_{6}^{1}\cdots
\times_{2m}^{1}\mathcal{G}_m
\Bigg)
\\[-1mm]
&\times_{(1,2m+2)}^{(N-m+3,2)}
\Bigg[
\mathcal{C}_2
\times_{(2,3,\ldots,N-m+1)}^{(3,5,\ldots,2(N-m)+1)}
\Bigg(
\mathcal{G}_{m+1}
\times_{4}^{1}\mathcal{G}_{m+2}
\times_{6}^{1}\cdots
\times_{2(N-m)}^{1}\mathcal{G}_N
\Bigg)
\Bigg].
\end{aligned}
\label{eq:c1_environment_contraction}
$$
An analogous equation holds for the second central core $\mathcal{C}_2$.
\end{theorem}
The subwheel equations in Theorem~\ref{the1} provide a unified basis for deriving alternating updates for all DCTW factors. With the other factors fixed, each outer-ring factor or central core enters its subproblem linearly, reducing the original nonlinear problem to a sequence of tractable linear subproblems.
%%%%%%%%%%%%%%%%%%%%%%%%%%%%%%%%%%%%%%%%%%%%%

\section{Model and Algorithm}\label{sect_3}
In this section, we apply the DCTW decomposition in Definition~\ref{def:dctw} to tensor completion and robust tensor principal component analysis and formulate the corresponding optimization models. Both models represent the underlying low-rank tensor using DCTW and are solved within the PAM framework, addressing missing data and sparse noise, respectively. Their convergence analysis follows the Kurdyka--{\L}ojasiewicz (KL) framework \cite{bolte2014proximal, xie2026accelerated, zheng2022tensor}. For brevity, we omit the proofs and focus on the computational complexity of the tensor completion algorithm.
\subsection{Tensor completion}\label{subsect3.1}
Assume $\mathcal{Y},\mathcal{X}\in\mathbb{R}^{I_1\times I_2\times\cdots\times I_N}$ denote the partially observed tensor and its underlying target tensor, respectively. Using the proposed dual-core tensor wheel decomposition, we formulate the tensor completion problem as
\begin{equation}
\label{eq_DCTW}
\begin{aligned}
\min_{\mathcal{X},\,\{\mathcal{G}_k\}_{k=1}^{N},
      \,\mathcal{C}_1,\,\mathcal{C}_2}
&\quad
\frac{1}{2}
\left\|
\mathcal{X}-
\operatorname{DCTW}
\left(
\{\mathcal{G}_k\}_{k=1}^{N};
\mathcal{C}_1,\mathcal{C}_2
\right)
\right\|_{F}^{2},
\\
\mathrm{s.t.}
&\quad
\mathcal{P}_{\Omega}
\left(\mathcal{X}-\mathcal{Y}\right)=0.
\end{aligned}
\end{equation}
where
$\mathcal{G}_k\in
\mathbb{R}^{R_k\times I_k\times L_k\times R_{k+1}}$
denotes the $k$th outer-ring factor, with $R_{N+1}=R_1$.
Here, $R_k$ is the ring rank and $L_k$ is the radial rank.
Let $m=\lceil N/2\rceil$. The two central cores are
$\mathcal{C}_1\in\mathbb{R}^{L_1\times\cdots\times L_m\times Q}$
and
$\mathcal{C}_2\in\mathbb{R}^{Q\times L_{m+1}\times\cdots\times L_N}$,
where $Q$ denotes their shared-bond rank. The corresponding mode groups are
$\mathbf{s}_1=(1,\ldots,m)$ and
$\mathbf{s}_2=(m+1,\ldots,N)$.
The observed index set is denoted by $\Omega$, and
$\mathcal{P}_{\Omega}(\cdot)$ is the associated sampling operator.
Problem~(\ref{eq_DCTW}) can be equivalently reformulated as the following unconstrained optimization problem
\begin{equation}
\label{eq_DCTW_unconstrained}
\begin{aligned}
\min_{\mathcal{X},\,\{\mathcal{G}_k\}_{k=1}^{N},
      \,\mathcal{C}_1,\,\mathcal{C}_2}
\quad
&
\frac{1}{2}
\left\|
\mathcal{X}
-
\operatorname{DCTW}
\left(
\{\mathcal{G}_k\}_{k=1}^{N};
\mathcal{C}_1,\mathcal{C}_2
\right)
\right\|_{F}^{2}
+
\iota_{\mathbb{D}}(\mathcal{X}),
\end{aligned}
\end{equation}
where
$$\label{eq_indicator_D}
\iota_{\mathbb{D}}(\mathcal{X})
:=
\begin{cases}
0, & \text{if } \mathcal{X}\in\mathbb{D},\\
+\infty, & \text{otherwise},
\end{cases} \text{where }\mathbb{D}
:=
\left\{
\mathcal{X}\in
\mathbb{R}^{I_1\times I_2\times\cdots\times I_N}
:
\mathcal{P}_{\Omega}
\left(\mathcal{X}-\mathcal{Y}\right)=0
\right\}.$$

The objective in~\eqref{eq_DCTW_unconstrained} is convex with respect to each variable block when the other blocks are fixed. We therefore apply proximal alternating minimization to sequentially update the outer-ring factors, central cores, and completed tensor. At iteration $t+1$, the updates are given by
\begin{equation}
\left\{
\begin{aligned}
\mathcal{G}_k^{(t+1)}
\in \arg\min_{\mathcal{G}_k}\quad
&
\frac{1}{2}
\left\|
\mathcal{X}^{(t)}
-
\operatorname{DCTW}
\left(
\mathcal{G}_{1:k-1}^{(t+1)},
\mathcal{G}_k,
\mathcal{G}_{k+1:N}^{(t)},
\mathcal{C}_1^{(t)},
\mathcal{C}_2^{(t)}
\right)
\right\|_F^2
+
\frac{\rho}{2}
\left\|
\mathcal{G}_k-\mathcal{G}_k^{(t)}
\right\|_F^2,
\\
\mathcal{C}_1^{(t+1)}
\in \arg\min_{\mathcal{C}_1}\quad
&
\frac{1}{2}
\left\|
\mathcal{X}^{(t)}
-
\operatorname{DCTW}
\left(
\left\{\mathcal{G}_k^{(t+1)}\right\}_{k=1}^{N},
\mathcal{C}_1,
\mathcal{C}_2^{(t)}
\right)
\right\|_F^2
+
\frac{\rho}{2}
\left\|
\mathcal{C}_1-\mathcal{C}_1^{(t)}
\right\|_F^2,
\\
\mathcal{C}_2^{(t+1)}
\in \arg\min_{\mathcal{C}_2}\quad
&
\frac{1}{2}
\left\|
\mathcal{X}^{(t)}
-
\operatorname{DCTW}
\left(
\left\{\mathcal{G}_k^{(t+1)}\right\}_{k=1}^{N},
\mathcal{C}_1^{(t+1)},
\mathcal{C}_2
\right)
\right\|_F^2
+
\frac{\rho}{2}
\left\|
\mathcal{C}_2-\mathcal{C}_2^{(t)}
\right\|_F^2,
\\
\mathcal{X}^{(t+1)}
\in \arg\min_{\mathcal{X}}\quad
&
\frac{1}{2}
\left\|
\mathcal{X}
-
\operatorname{DCTW}
\left(
\left\{\mathcal{G}_k^{(t+1)}\right\}_{k=1}^{N},
\mathcal{C}_1^{(t+1)},
\mathcal{C}_2^{(t+1)}
\right)
\right\|_F^2
+
\frac{\rho}{2}
\left\|
\mathcal{X}-\mathcal{X}^{(t)}
\right\|_F^2
+
\iota_{\mathbb{D}}(\mathcal{X}),
\end{aligned}
\right.
\label{eq_DCTW_PAM}
\end{equation}
where $k=1,\ldots,N$, $\rho>0$ is the proximal parameter, and $t$ is the iteration index.

\textbf{(1) Updating the outer-ring factors
$\mathcal{G}_k$ for $k=1,2,\ldots,N$:}
When updating $\mathcal{G}_k$, we fix
$\{\mathcal{G}_i^{(t+1)}\}_{i<k}$ at their updated values and
$\{\mathcal{G}_i^{(t)}\}_{i>k}$, $\mathcal{C}_1^{(t)}$, and
$\mathcal{C}_2^{(t)}$ at their current values. Contracting all fixed
factors except $\mathcal{G}_k$ yields the environment matrix
$\mathbf{M}_{\neq k}^{(t)}$, as established in
Theorem~\ref{the1}. The $\mathcal{G}_k$-subproblem
in~\eqref{eq_DCTW_PAM} can therefore be written as
$$
\begin{aligned}
\mathbf{G}_{k(2)}^{(t+1)}
\in
\arg\min_{\mathbf{G}_{k(2)}}\quad
&
\frac{1}{2}
\left\|
\mathbf{X}_{\langle k\rangle}^{(t)}
-
\mathbf{G}_{k(2)}\mathbf{M}_{\neq k}^{(t)}
\right\|_F^2
+
\frac{\rho}{2}
\left\|
\mathbf{G}_{k(2)}
-
\mathbf{G}_{k(2)}^{(t)}
\right\|_F^2.
\end{aligned}
$$
Consequently, the closed-form solution is
\begin{equation}
\label{eq_DCTW_gk}
\begin{aligned}
\mathcal{G}_k^{(t+1)}
=
\operatorname{Fold}_2
\Bigg(
&
\left(
\mathbf{X}_{\langle k\rangle}^{(t)}
\left(\mathbf{M}_{\neq k}^{(t)}\right)^{\mathrm T}
+
\rho\mathbf{G}_{k(2)}^{(t)}
\right)
\left(
\mathbf{M}_{\neq k}^{(t)}
\left(\mathbf{M}_{\neq k}^{(t)}\right)^{\mathrm T}
+
\rho\mathbf{I}
\right)^{-1}
\Bigg).
\end{aligned}
\end{equation}

\textbf{(2) Updating the central cores
$\mathcal{C}_1$ and $\mathcal{C}_2$:}
After updating all outer-ring factors, we update the two central cores
sequentially. We first update $\mathcal{C}_1$ while fixing
$\mathcal{C}_2=\mathcal{C}_2^{(t)}$. The updated core
$\mathcal{C}_1^{(t+1)}$ is then used to update $\mathcal{C}_2$.
The $\mathcal{C}_1$-subproblem in~\eqref{eq_DCTW_PAM} is
$$
\mathbf{c}_1^{(t+1)}
\in
\arg\min_{\mathbf{c}_1}
\frac{1}{2}
\left\|
\mathbf{x}^{(t)}
-
\mathbf{c}_1
\mathbf{N}_{\neq\mathcal{C}_1}^{(t)}
\right\|_F^2
+
\frac{\rho}{2}
\left\|
\mathbf{c}_1-\mathbf{c}_1^{(t)}
\right\|_F^2.
$$
Its closed-form solution is
\begin{equation}
\mathcal{C}_1^{(t+1)}
=
\operatorname{GFold}
\left(
\left(
\mathbf{x}^{(t)}
\left(\mathbf{N}_{\neq\mathcal{C}_1}^{(t)}\right)^{\mathsf T}
+
\rho\mathbf{c}_1^{(t)}
\right)
\left(
\mathbf{N}_{\neq\mathcal{C}_1}^{(t)}
\left(\mathbf{N}_{\neq\mathcal{C}_1}^{(t)}\right)^{\mathsf T}
+
\rho\mathbf{I}
\right)^{-1},
\varnothing\mid1:m+1
\right).
\label{eq:c1_tensor_closed_form}
\end{equation}
The $\mathcal{C}_2$-subproblem has the same structure, yielding
\begin{equation}
\mathcal{C}_2^{(t)}
=
\operatorname{GFold}
\left(
\left[
\mathbf{x}^{(t)}
\left(
\mathbf{N}_{\neq\mathcal{C}_2}^{(t)}
\right)^{\mathsf T}
+
\rho\mathbf{c}_2^{(t)}
\right]
\left[
\mathbf{N}_{\neq\mathcal{C}_2}^{(t)}
\left(
\mathbf{N}_{\neq\mathcal{C}_2}^{(t)}
\right)^{\mathsf T}
+
\rho\mathbf{I}
\right]^{-1},
\varnothing\mid1:N-m+1
\right).
\label{eq:c2_tensor_closed_form}
\end{equation}

\textbf{(3) Updating the completed tensor $\mathcal{X}$:}
The $\mathcal{X}$-subproblem in~\eqref{eq_DCTW_PAM} can likewise be solved in closed form 
\begin{equation}
\mathcal{X}^{(t+1)}
=
\mathcal{P}_{\Omega^c}
\left(
\frac{
\operatorname{DCTW}
\left(
\{\mathcal{G}_k^{(t+1)}\}_{k=1}^{N},
\mathcal{C}_1^{(t+1)},
\mathcal{C}_2^{(t+1)}
\right)
+
\rho\mathcal{X}^{(t)}
}{
1+\rho
}
\right)
+
\mathcal{P}_{\Omega}(\mathcal{Y}),
\label{eq_DCTW_30}
\end{equation}
where $\Omega^c$ denotes the complement of the observed index set
$\Omega$. We summarize the resulting PAM-based optimization procedure for DCTW in Algorithm~\ref{alg:DCTW_TC_PAM}.

\begin{algorithm}[!ht]
\caption{PAM-based solver for DCTW tensor completion}
\label{alg:DCTW_TC_PAM}

\noindent\textbf{Input:}\enspace
The observed tensor
$\mathcal{Y}\in\mathbb{R}^{I_1\times\cdots\times I_N}$,
the observed index set $\Omega$, the DCTW rank
$\mathbf{r}_{\mathrm{DCTW}}=(\mathbf{R},\mathbf{L},Q)$,
the proximal parameter $\rho=0.1$, the maximum number of iterations
$t_{\max}=500$, and the tolerance $\varepsilon=10^{-5}$.

\par\smallskip
\noindent\textbf{Initialization:}
Set $t=0$ and $\mathcal{X}^{(0)}=\mathcal{Y}$.
Randomly initialize
$\mathcal{G}_k^{(0)}
\in\mathbb{R}^{R_k\times I_k\times L_k\times R_{k+1}}$,
$k=1,\ldots,N$, with $R_{N+1}=R_1$, and
$\mathcal{C}_1^{(0)}
\in\mathbb{R}^{L_1\times\cdots\times L_m\times Q}$
and
$\mathcal{C}_2^{(0)}
\in\mathbb{R}^{Q\times L_{m+1}\times\cdots\times L_N}$,
where $m=\lceil N/2\rceil$.

\par\smallskip
\begin{algorithmic}[1]
\While{not converged and $t<t_{\max}$}
    \State Update $\mathcal{G}_k^{(t+1)}$, for
    $k=1,\ldots,N$, using \eqref{eq_DCTW_gk}. 
    \State Update $\mathcal{C}_1^{(t+1)}$ using
    \eqref{eq:c1_tensor_closed_form}. 
    \State Update $\mathcal{C}_2^{(t+1)}$ using
    \eqref{eq:c2_tensor_closed_form}. \textcolor{blue}{$\leftarrow$ Accelerated updates for steps 2--4 (section~\ref{subsect4.2})}
    \State Update $\mathcal{X}^{(t+1)}$ using
    \eqref{eq_DCTW_30}.
    \State Check whether the stopping criterion: 
    $\left\|\mathcal{X}^{(t+1)}-\mathcal{X}^{(t)}\right\|_F/
    \left\|\mathcal{X}^{(t)}\right\|_F<\varepsilon$.
    \State $t\leftarrow t+1$.
\EndWhile
\end{algorithmic}

\par\smallskip
\noindent\textbf{Output:}\enspace
The recovered tensor
$\mathcal{X}\in\mathbb{R}^{I_1\times\cdots\times I_N}$.
\end{algorithm}

% \noindent\textbf{Computational complexity.}
% The computational cost of the Algorithm \ref{alg:DCTW_TC_PAM} mainly arises
% from updating the outer ring factors $\{\mathcal{G}_k\}_{k=1}^{N}$,
% the two central cores $\mathcal{C}_1$ and $\mathcal{C}_2$, and the
% completed tensor $\mathcal{X}$. For compactness, define
% \[
% P_I=\prod_{k=1}^{N}I_k,\qquad
% P_L=\prod_{k=1}^{N}L_k,\qquad
% P_{L,1}=\prod_{k=1}^{m}L_k,\qquad
% P_{L,2}=\prod_{k=m+1}^{N}L_k,
% \]
% where $m=\lceil N/2\rceil$ and
% $d_k=R_kR_{k+1}L_k$ with $R_{N+1}=R_1$.
% The costs of updating the outer ring factors, the two central cores,
% and the completed tensor are, respectively,
% $$
% \begin{aligned}
% \mathcal{C}_{\mathcal{G}}
% &=
% \mathcal{O}\left(
% \sum_{k=1}^{N}
% \left(
% P_Id_k+\frac{P_I}{I_k}d_k^2+d_k^3
% \right)
% \right),                                                   
% \mathcal{C}_{\mathcal{C}}
% =
% \mathcal{O}\left(
% Q^2P_I\left(P_{L,1}^2+P_{L,2}^2\right)
% +
% Q^3\left(P_{L,1}^3+P_{L,2}^3\right)
% \right),                                                   \\
% \mathcal{C}_{\mathcal{X}}
% &=
% \mathcal{O}\left(R_1R_NP_IP_L\right).
% \end{aligned}
% $$
% Therefore, assuming that $R_k=R$ and
% $I_k\geq\max\{R^2,N\}$, the computational complexity of one complete
% PAM iteration can be simplified as

% $$\mathcal{C}_{\mathrm{iter}}
% =
% \mathcal{O}\left(
% R^2P_IP_L
% +
% Q^2P_I\left(P_{L,1}^2+P_{L,2}^2\right)
% +
% Q^3\left(P_{L,1}^3+P_{L,2}^3\right)
% \right).$$

% Required package: \usepackage{amsmath}

\textbf{Computational complexity.}
We analyze the computational complexity of
Algorithm~\ref{alg:DCTW_TC_PAM} using unaccelerated explicit updates.
The main computational costs arise from updating the outer-ring factors
$\{\mathcal{G}_k\}_{k=1}^{N}$, the central cores
$\mathcal{C}_1$ and $\mathcal{C}_2$, and the completed tensor
$\mathcal{X}$. Define
$P_I=\prod_{k=1}^{N}I_k$,
$P_L=\prod_{k=1}^{N}L_k$,
$d_k=R_kR_{k+1}L_k$,
$A=\prod_{k=1}^{m}L_k$,
$B=\prod_{k=m+1}^{N}L_k$, and
$m=\lceil N/2\rceil$, where $R_{N+1}=R_1$. We also define
$a_{k,j}=1+\operatorname{mod}(k+j-1,N)$.
The cost of constructing the complete outer-ring environment is
$$
S=
\sum_{i=2}^{N-1}
R_1R_iR_{i+1}\prod_{j=1}^{i}I_jL_j
+
R_1R_NP_IP_L.
$$
The total cost of updating all $N$ outer-ring factors is
\begin{equation}
\begin{aligned}
\mathcal{C}_{\mathcal{G}}
=
\mathcal{O}\Biggl(
&\sum_{k=1}^{N}
\left(
QP_L+\frac{P_IP_L}{I_k}R_kR_{k+1}
\right)
+
\sum_{k=1}^{N}\sum_{j=2}^{N-1}
\left(
R_{k+1}R_{a_{k,j}}R_{a_{k,j}+1}
\prod_{h=1}^{j}I_{a_{k,h}}L_{a_{k,h}}
\right)
\\
&+
\sum_{k=1}^{N}P_Id_k
+
\sum_{k=1}^{N}\frac{P_I}{I_k}d_k^2
+
\sum_{k=1}^{N}\left(d_k^3+I_kd_k^2\right)
\Biggr).
\end{aligned}
\end{equation}
The costs of updating the two central cores and $\mathcal{X}$ are
$$
\begin{aligned}
\mathcal{C}_{\mathcal{C}_1}
&=
\mathcal{O}\left(
S+QP_IP_L+P_IQA+P_IQ^2A^2+Q^3A^3
\right),
\\
\mathcal{C}_{\mathcal{C}_2}
&=
\mathcal{O}\left(
S+QP_IP_L+P_IQB+P_IQ^2B^2+Q^3B^3
\right),
\\
\mathcal{C}_{\mathcal{X}}
&=
\mathcal{O}\left(
QP_L+S+P_IP_L+P_I
\right).
\end{aligned}
$$
Consequently, the total computational complexity is
$$
\mathcal{C}
=
\mathcal{C}_{\mathcal{G}}
+
\mathcal{C}_{\mathcal{C}_1}
+
\mathcal{C}_{\mathcal{C}_2}
+
\mathcal{C}_{\mathcal{X}}.
$$
The proposed acceleration strategy reduces the costs represented by
the first three terms of
$\mathcal{C}_{\mathcal{G}}$,
$\mathcal{C}_{\mathcal{C}_1}$, and
$\mathcal{C}_{\mathcal{C}_2}$ by caching and reusing intermediate
contractions during the construction and contraction of the
environment tensors.

\subsection{Tensor robust principal component analysis}
Tensor robust principal component analysis (TRPCA) decomposes a
high-order tensor into low-rank and sparse components. This decomposition
is used for sparse-noise removal and anomaly detection in high-dimensional
data. Let the observed tensor
$\mathcal{Y}\in\mathbb{R}^{I_1\times\cdots\times I_N}$ satisfy
$$
\mathcal{Y}=\mathcal{X}+\mathcal{E},
$$
where $\mathcal{X}\in\mathbb{R}^{I_1\times\cdots\times I_N}$ is the
underlying low-rank component and
$\mathcal{E}\in\mathbb{R}^{I_1\times\cdots\times I_N}$ is the sparse
component. The DCTW based TRPCA model is formulated as
\begin{equation}
\label{eq:DCTW_TRPCA}
\begin{aligned}
\min_{\mathcal{X},\mathcal{E},
\{\mathcal{G}_k\}_{k=1}^{N},\mathcal{C}_1,\mathcal{C}_2}\quad
&
\frac{1}{2}
\left\|
\mathcal{X}
-
\operatorname{DCTW}
\left(
\{\mathcal{G}_k\}_{k=1}^{N};
\mathcal{C}_1,\mathcal{C}_2
\right)
\right\|_F^2
+
\alpha\|\mathcal{E}\|_1
\\
\mathrm{s.t.}\quad
&
\mathcal{Y}=\mathcal{X}+\mathcal{E},
\end{aligned}
\end{equation}
where $\alpha>0$ controls the trade-off between the low-rank
approximation and the sparsity of $\mathcal{E}$. Adding the quadratic
penalty term
$\frac{\eta}{2}\|\mathcal{Y}-(\mathcal{X}+\mathcal{E})\|_F^2$
yields the following penalized formulation of
model~\eqref{eq:DCTW_TRPCA}:
\begin{equation}
\label{eq:DCTW_TRPCA_penalized}
\begin{aligned}
\min_{\mathcal{X},\mathcal{E},
\{\mathcal{G}_k\}_{k=1}^{N},\mathcal{C}_1,\mathcal{C}_2}\quad
&
\frac{1}{2}
\left\|
\mathcal{X}
-
\operatorname{DCTW}
\left(
\{\mathcal{G}_k\}_{k=1}^{N};
\mathcal{C}_1,\mathcal{C}_2
\right)
\right\|_F^2
+
\frac{\eta}{2}
\left\|
\mathcal{Y}-\mathcal{X}-\mathcal{E}
\right\|_F^2
+
\alpha\|\mathcal{E}\|_1,
\end{aligned}
\end{equation}
where $\eta>0$ is the penalty parameter.

We solve~\eqref{eq:DCTW_TRPCA_penalized} using PAM. The outer-ring
factors $\{\mathcal{G}_k\}_{k=1}^{N}$ and the central cores
$\mathcal{C}_1$ and $\mathcal{C}_2$ are updated using the same scheme
described in Section~\ref{subsect3.1}. After these updates, the sparse
component $\mathcal{E}$ and the low-rank component $\mathcal{X}$ are
updated sequentially by solving their corresponding subproblems.

The $\mathcal{E}$-subproblem is
$$
\mathcal{E}^{(t+1)}
\in
\arg\min_{\mathcal{E}}
\frac{\eta}{2}
\left\|
\mathcal{Y}-\mathcal{X}^{(t)}-\mathcal{E}
\right\|_F^2
+
\alpha\|\mathcal{E}\|_1
+
\frac{\rho}{2}
\left\|
\mathcal{E}-\mathcal{E}^{(t)}
\right\|_F^2,
$$
where $\rho>0$ is the proximal parameter. Its closed-form solution is
\begin{equation}
\label{eq:DCTW_TRPCA_E}
\mathcal{E}^{(t+1)}
=
\mathcal{S}_{\frac{\alpha}{\eta+\rho}}
\left(
\frac{
\eta\left(\mathcal{Y}-\mathcal{X}^{(t)}\right)
+
\rho\mathcal{E}^{(t)}
}{
\eta+\rho
}
\right),
\end{equation}
where
$\mathcal{S}_{\tau}(z)=\operatorname{sign}(z)
\max\{|z|-\tau,0\}$
is the elementwise soft-thresholding operator.

Let
$\mathcal{Z}^{(t+1)}=
\operatorname{DCTW}\left(
\{\mathcal{G}_k^{(t+1)}\}_{k=1}^{N};
\mathcal{C}_1^{(t+1)},\mathcal{C}_2^{(t+1)}
\right)$
denote the current DCTW reconstruction. The $\mathcal{X}$-subproblem is
\begin{equation}
\label{eq:DCTW_TRPCA_X_subproblem}
\begin{aligned}
\mathcal{X}^{(t+1)}
\in\arg\min_{\mathcal{X}}\quad
&
\frac{1}{2}
\left\|
\mathcal{X}-\mathcal{Z}^{(t+1)}
\right\|_F^2
+
\frac{\eta}{2}
\left\|
\mathcal{Y}-\mathcal{X}-\mathcal{E}^{(t+1)}
\right\|_F^2
+
\frac{\rho}{2}
\left\|
\mathcal{X}-\mathcal{X}^{(t)}
\right\|_F^2.
\end{aligned}
\end{equation}
The closed-form solution to~\eqref{eq:DCTW_TRPCA_X_subproblem} is
\begin{equation}
\label{eq:DCTW_TRPCA_X}
\mathcal{X}^{(t+1)}
=
\frac{
\mathcal{Z}^{(t+1)}
+
\eta\left(\mathcal{Y}-\mathcal{E}^{(t+1)}\right)
+
\rho\mathcal{X}^{(t)}
}{
1+\eta+\rho
}.
\end{equation}
The PAM procedure for DCTW based TRPCA is summarized in
Algorithm~\ref{alg:DCTW_TRPCA_PAM}.

\begin{algorithm}[!ht]
\caption{PAM-based solver for DCTW denoising}
\label{alg:DCTW_TRPCA_PAM}

\noindent\textbf{Input:}\enspace
The noisy tensor
$\mathcal{Y}\in\mathbb{R}^{I_1\times\cdots\times I_N}$,
the DCTW rank
$\mathbf{r}_{\mathrm{DCTW}}=(\mathbf{R},\mathbf{L},Q)$,
the noise regularization parameter $\alpha \in[0.01,0.001]$, the penalty parameter
$\eta$, the proximal parameter $\rho=0.1$, the maximum number of
iterations $t_{\max} \in \{700,800,900\}$, and the tolerance
$\varepsilon=10^{-5}$.

\par\smallskip
\noindent\textbf{Initialization:}\enspace
Set $t=0$, $\mathcal{X}^{(0)}=\mathcal{Y}$, and
$\mathcal{E}^{(0)}=\mathbf{0}$.
Randomly initialize
$\mathcal{G}_k^{(0)}
\in\mathbb{R}^{R_k\times I_k\times L_k\times R_{k+1}}$,
$k=1,\ldots,N$, with $R_{N+1}=R_1$, and
$\mathcal{C}_1^{(0)}
\in\mathbb{R}^{L_1\times\cdots\times L_m\times Q}$
and
$\mathcal{C}_2^{(0)}
\in\mathbb{R}^{Q\times L_{m+1}\times\cdots\times L_N}$,
where $m=\lceil N/2\rceil$.

\par\smallskip
\begin{algorithmic}[1]
\While{not converged and $t<t_{\max}$}
    \State Update $\mathcal{G}_k^{(t+1)}$, for
    $k=1,\ldots,N$, using \eqref{eq_DCTW_gk}.
     \State Update $\mathcal{C}_1^{(t+1)}$ using
    \eqref{eq:c1_tensor_closed_form}.
    \State Update $\mathcal{C}_2^{(t+1)}$ using
    \eqref{eq:c2_tensor_closed_form}. \textcolor{blue}{$\leftarrow$ Accelerated updates for steps 2--4 (section~\ref{subsect4.2})}
    \State Update $\mathcal{E}^{(t+1)}$ using
    \eqref{eq:DCTW_TRPCA_E}.
    \State Update $\mathcal{X}^{(t+1)}$ using
    \eqref{eq:DCTW_TRPCA_X}.
    \State Check whether the stopping criterion is satisfied:
    $\left\|\mathcal{X}^{(t+1)}-\mathcal{X}^{(t)}\right\|_F/
    \left\|\mathcal{X}^{(t)}\right\|_F<\varepsilon$.
    \State $t\leftarrow t+1$.
\EndWhile
\end{algorithmic}

\par\smallskip
\noindent\textbf{Output:}\enspace
The denoised tensor
$\mathcal{X}\in\mathbb{R}^{I_1\times\cdots\times I_N}$.
\end{algorithm}

\section{Implicit Updating and Structural Acceleration}\label{sect_4}
The PAM algorithms developed in the preceding section require repeated matrix products involving large environment matrices. For high-dimensional tensors, explicitly constructing these matrices leads to substantial computational and memory costs. This section introduces two acceleration strategies. First, we evaluate the matrix products through generalized tensor contractions, thereby avoiding the explicit formation of environment matrices. Second, we develop a structure-cached strategy that reuses intermediate contractions across block updates. These strategies reduce redundant computation and memory use while producing results equivalent to those of the explicit updates, and can be extended to other tensor network algorithms that require the explicit construction of large matrices and tensor contractions.

\subsection{Implicit updating}
Implicit updating has been studied in
\cite{zdunek2026fasttwd}. We briefly review this strategy
to establish the basis for the structure-cached strategy introduced in
Section~\ref{subsect4.2}, which extends implicit updating by reusing
intermediate contractions.

For clarity, we first consider a fourth-order tensor and then extend the
analysis to an $N$th-order tensor. For a fourth-order tensor, the $k$th
outer-ring factor, where $k\in\{1,2,3,4\}$, has the following closed-form
update
\begin{equation}
\label{eq_DCTW_gk_fourth}
\begin{aligned}
\mathcal{G}_k^{(t+1)}
=
\operatorname{Fold}_2
\Bigg(
&
\left(
\mathbf{X}_{\langle k\rangle}^{(t)}
\left(\mathbf{M}_{\neq k}^{(t)}\right)^{\mathrm T}
+
\rho \mathbf{G}_{k(2)}^{(t)}
\right)
\left(
\mathbf{M}_{\neq k}^{(t)}
\left(\mathbf{M}_{\neq k}^{(t)}\right)^{\mathrm T}
+
\rho\mathbf{I}
\right)^{-1}
\Bigg),
\end{aligned}
\end{equation}
The dominant computational costs in~\eqref{eq_DCTW_gk_fourth} arise from
the matrix products
$\mathbf{X}_{\langle k\rangle}^{(t)}
\left(\mathbf{M}_{\neq k}^{(t)}\right)^{\mathrm T}$
and
$\mathbf{M}_{\neq k}^{(t)}
\left(\mathbf{M}_{\neq k}^{(t)}\right)^{\mathrm T}$.
Their direct evaluation requires constructing the subwheel tensor
$\mathcal{M}_{\neq k}^{(t)}$ through generalized tensor contractions,
unfolding it into a matrix, and performing the resulting matrix
multiplications. Because $\mathcal{M}_{\neq k}^{(t)}$ is typically large,
its explicit construction and unfolding make these operations
computationally expensive.

To avoid explicitly constructing the subwheel tensor
$\mathcal{M}_{\neq k}^{(t)}$, we compute the data term through tensor
contractions. After an appropriate mode permutation,
$\mathcal{X}^{(t)}$ is sequentially contracted with the outer-ring
factors $\mathcal{G}_j^{(t)}$, where
$j\in\{1,\ldots,N\}\setminus\{k\}$, and the central cores
$\mathcal{C}_1^{(t)}$ and $\mathcal{C}_2^{(t)}$. Generalized unfolding
of the resulting tensor directly yields
$\mathbf{X}_{\langle k\rangle}^{(t)}
\left(\mathbf{M}_{\neq k}^{(t)}\right)^{\mathrm T}$.
This implicit formulation avoids constructing and unfolding the
subwheel tensor and eliminates the associated large-scale matrix
multiplication.

For the fourth-order case, we use the update of
$\mathcal{G}_1^{(t)}$ as an example. The corresponding contraction is
\begin{equation}
\label{eq_implicit_data_term_G1}
\begin{aligned}
&\mathbf{X}_{(1)}^{(t)}
\left(\mathbf{M}_{\neq 1}^{(t)}\right)^{\mathrm T}
=
\operatorname{Unfold}_1
\Bigg[
\Bigg(
\Bigg(
\Big(
\mathcal{X}^{(t)}
\times_{2}^{2}
\mathcal{G}_2^{(t)}
\Big)
\times_{2,6}^{1,2}
\mathcal{G}_3^{(t)}
\Bigg)
\times_{2,6}^{1,2}
\mathcal{G}_4^{(t)}
\Bigg)
\times_{3,4,5}^{2,3,4}
\Big(
\mathcal{C}_1^{(t)}
\times_{2}^{2}
\mathcal{C}_2^{(t)}
\Big)
\Bigg].
\end{aligned}
\end{equation}
Similarly, the Gram term
$\mathbf{M}_{\neq 1}^{(t)}
\left(\mathbf{M}_{\neq 1}^{(t)}\right)^{\mathrm T}$
can be evaluated directly in tensor-contraction form:
\begin{equation}
\label{eq_implicit_gram_term_G1}
\begin{aligned}
&\mathbf{M}_{\neq 1}^{(t)}
\left(
\mathbf{M}_{\neq 1}^{(t)}
\right)^{\mathrm T}
=
\operatorname{GUnfold}
\Bigg(
\Bigg[
\left(
\mathcal{C}_1^{(t)}
\times_3^2
\left(
\mathcal{G}_2^{(t)}
\times_2^2
\mathcal{G}_2^{(t)}
\right)
\right)
\times_6^3
\mathcal{C}_1^{(t)}
\Bigg]
\\
&\times_{2,4,6,8}^{1,2,3,4}
\Bigg[
\left(
\left(
\mathcal{C}_2^{(t)}
\times_1^2
\left(
\mathcal{G}_3^{(t)}
\times_2^2
\mathcal{G}_3^{(t)}
\right)
\right)
\times_6^1
\mathcal{C}_2^{(t)}
\right)
\times_{2,4,6,8}^{2,1,4,5}
\left(
\mathcal{G}_4^{(t)}
\times_2^2
\mathcal{G}_4^{(t)}
\right)
\Bigg],
5,1,2
\mid
6,4,3
\Bigg).
\end{aligned}
\end{equation}
The contraction identities derived for the fourth-order case extend
directly to an $N$th-order tensor. Taking the update of
$\mathcal{G}_1$ as an example, we sequentially contract
$\mathcal{X}^{(t)}$ with the remaining outer-ring factors
$\{\mathcal{G}_j^{(t)}\}_{j=2}^{N}$ and the central cores
$\mathcal{C}_1^{(t)}$ and $\mathcal{C}_2^{(t)}$. This yields the
following implicit representation
$$
\begin{aligned}
\mathcal{Y}_2^{(t)}
&=
\mathcal{X}^{(t)}
\times_2^2
\mathcal{G}_2^{(t)},\quad
\mathcal{Y}_j^{(t)}
=
\mathcal{Y}_{j-1}^{(t)}
\times_{2,N+2}^{1,2}
\mathcal{G}_j^{(t)},
\qquad j=3,\ldots,N.
\end{aligned}
$$
The two central cores are contracted along their shared bond of
dimension $Q$, yielding
$
\mathcal{H}^{(t)}
=
\mathcal{C}_1^{(t)}
\times_{m+1}^{1}
\mathcal{C}_2^{(t)}.
$
We then contract $\mathcal{Y}_N^{(t)}$ with
$\mathcal{H}^{(t)}$ along the radial modes
$L_2,\ldots,L_N$. Generalized unfolding of the resulting tensor gives
the data term
$$
\begin{aligned}
\mathcal{A}_1^{(t)}
&=
\mathcal{Y}_N^{(t)}
\times_{3,4,\ldots,N+1}^{2,3,\ldots,N}
\mathcal{H}^{(t)},\quad
\mathbf{X}_{(1)}^{(t)}
\left(
\mathbf{M}_{\neq1}^{(t)}
\right)^{\mathrm T}
=
\operatorname{GUnfold}
\left(
\mathcal{A}_1^{(t)},
1\mid2,3,4
\right).
\end{aligned}
$$
The same construction applies to any outer-ring factor
$\mathcal{G}_k$, $k=1,\ldots,N$, after a cyclic permutation of the
tensor modes.

For the Gram term associated with the update of $\mathcal{G}_1$, we
introduce an identical copy
$\widehat{\mathcal{G}}_j^{(t)}=\mathcal{G}_j^{(t)}$. The hats
distinguish the modes of the two copies. Contracting each pair along
its physical mode gives
$$
\mathcal{T}_j^{(t)}
=
\mathcal{G}_j^{(t)}
\times_2^2
\widehat{\mathcal{G}}_j^{(t)},
\qquad j=2,\ldots,N,
$$
with mode ordering
$$
\mathcal{T}_j^{(t)}
:
\left(
R_j,L_j,R_{j+1},
\widehat{R}_j,\widehat{L}_j,\widehat{R}_{j+1}
\right).
$$
We similarly introduce copies
$\widehat{\mathcal{C}}_1^{(t)}=\mathcal{C}_1^{(t)}$ and
$\widehat{\mathcal{C}}_2^{(t)}=\mathcal{C}_2^{(t)}$ of the central
cores. According to
Definition~\ref{def:dimension_set_contraction}, the contraction
dimension sets for the left and right Gram branches are
$$
\begin{aligned}
\Omega_{\mathrm L}^{(1)}
={}&
\bigcup_{j=2}^{m}
\left\{
\left(
L_j^{(\mathcal{T}_j)},
L_j^{(\mathcal{C}_1)}
\right),
\left(
\widehat{L}_j^{(\mathcal{T}_j)},
\widehat{L}_j^{(\widehat{\mathcal{C}}_1)}
\right)
\right\}
\cup
\bigcup_{j=2}^{m-1}
\left\{
\left(
R_{j+1}^{(\mathcal{T}_j)},
R_{j+1}^{(\mathcal{T}_{j+1})}
\right),
\left(
\widehat{R}_{j+1}^{(\mathcal{T}_j)},
\widehat{R}_{j+1}^{(\mathcal{T}_{j+1})}
\right)
\right\},
\\
\Omega_{\mathrm R}^{(1)}
={}&
\bigcup_{j=m+1}^{N}
\left\{
\left(
L_j^{(\mathcal{T}_j)},
L_j^{(\mathcal{C}_2)}
\right),
\left(
\widehat{L}_j^{(\mathcal{T}_j)},
\widehat{L}_j^{(\widehat{\mathcal{C}}_2)}
\right)
\right\}
\cup
\bigcup_{j=m+1}^{N-1}
\left\{
\left(
R_{j+1}^{(\mathcal{T}_j)},
R_{j+1}^{(\mathcal{T}_{j+1})}
\right),
\left(
\widehat{R}_{j+1}^{(\mathcal{T}_j)},
\widehat{R}_{j+1}^{(\mathcal{T}_{j+1})}
\right)
\right\}.
\end{aligned}
$$
The left and right Gram branches are then computed as
$$
\begin{aligned}
\mathcal{L}_1
&=
\operatorname{contract}
\left(
\mathcal{C}_1^{(t)},
\widehat{\mathcal{C}}_1^{(t)},
\mathcal{T}_2^{(t)},\ldots,\mathcal{T}_m^{(t)};
\Omega_{\mathrm L}^{(1)}
\right), \quad
\mathcal{R}_1
=
\operatorname{contract}
\left(
\mathcal{C}_2^{(t)},
\widehat{\mathcal{C}}_2^{(t)},
\mathcal{T}_{m+1}^{(t)},\ldots,\mathcal{T}_N^{(t)};
\Omega_{\mathrm R}^{(1)}
\right),
\end{aligned}
$$
with retained mode orderings
$$
\begin{aligned}
\mathcal{L}_1
&:
\left(
L_1,Q,R_2,R_{m+1},
\widehat{R}_2,\widehat{R}_{m+1},
\widehat{L}_1,\widehat{Q}
\right), \quad
\mathcal{R}_1
:
\left(
Q,R_{m+1},\widehat{R}_{m+1},
\widehat{Q},R_1,\widehat{R}_1
\right).
\end{aligned}
$$
The two branches are connected along the shared central and boundary
ring modes using
$$
\Omega_{\mathrm{LR}}
=
\left\{
\left(
Q^{(\mathcal{L}_1)},
Q^{(\mathcal{R}_1)}
\right),
\left(
R_{m+1}^{(\mathcal{L}_1)},
R_{m+1}^{(\mathcal{R}_1)}
\right),
\left(
\widehat{R}_{m+1}^{(\mathcal{L}_1)},
\widehat{R}_{m+1}^{(\mathcal{R}_1)}
\right),
\left(
\widehat{Q}^{(\mathcal{L}_1)},
\widehat{Q}^{(\mathcal{R}_1)}
\right)
\right\}.
$$
The complete Gram tensor obtained after removing
$\mathcal{G}_1$ is
$$
\boldsymbol{\Gamma}_{\neq1}^{(t)}
=
\operatorname{contract}
\left(
\mathcal{L}_1,\mathcal{R}_1;
\Omega_{\mathrm{LR}}
\right),
$$
with mode ordering
$
\boldsymbol{\Gamma}_{\neq1}^{(t)}
:
\left(
L_1,R_2,\widehat{R}_2,
\widehat{L}_1,R_1,\widehat{R}_1
\right).
$
Using $(R_1,L_1,R_2)$ as the row modes and
$(\widehat{R}_1,\widehat{L}_1,\widehat{R}_2)$ as the column modes
gives
$$
\mathbf{M}_{\neq1}^{(t)}
\left(\mathbf{M}_{\neq1}^{(t)}\right)^{\mathrm T}
=
\operatorname{GUnfold}
\left(
\boldsymbol{\Gamma}_{\neq1}^{(t)},
5,1,2\mid6,4,3
\right).
$$
The same Gram-matrix construction applies to the other outer-ring
factors after adjusting the contraction dimension sets.

\subsection{Structural acceleration}\label{subsect4.2}
In this section, we extend the implicit update scheme by exploiting the structure of the dual-core tensor wheel decomposition. The resulting structure-cached strategy reuses intermediate contractions across block updates, reducing computation and memory use while preserving the original closed-form solutions and numerical results.

\begin{figure}[!ht]
    \centering
    \includegraphics[width=1.0\linewidth]{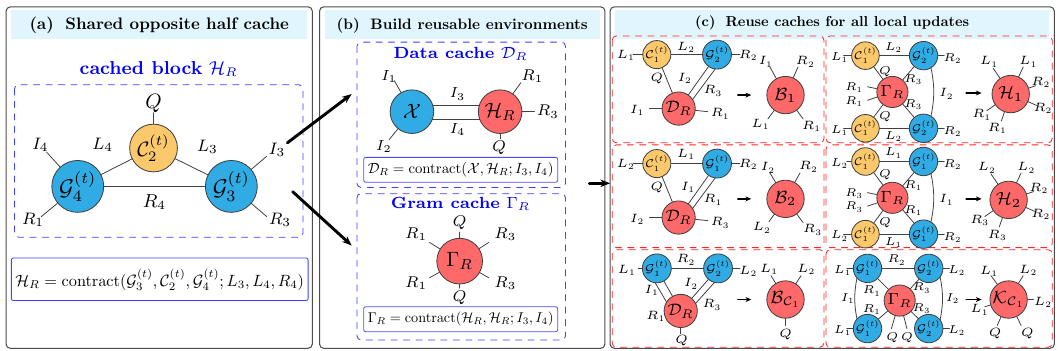}
    \caption{Structure-cached acceleration.}
    \label{fig_2}
\end{figure}

To illustrate the structure-cached acceleration strategy, we consider a
fourth-order tensor. Because this case involves only a few contracted
modes, we list the mode labels directly rather than specifying explicit
mode pairs. We first contract and cache the complementary right-half
subnetwork:
\begin{equation}
\label{eq_cached_HR}
\mathcal{H}_{\mathrm R}
=
\operatorname{contract}
\left(
\mathcal{G}_3^{(t)},
\mathcal{C}_2^{(t)},
\mathcal{G}_4^{(t)};
L_3,L_4,R_4
\right),
\end{equation}
where
$\mathcal{H}_{\mathrm R}
\in
\mathbb{R}^{R_3\times I_3\times I_4\times R_1\times Q}$.
We then contract $\mathcal{H}_{\mathrm R}$ with a copy of itself to
construct the Gram cache and with the data tensor $\mathcal{X}^{(t)}$
to construct the data cache:
\begin{equation}
\label{eq_cached_gramR}
\boldsymbol{\Gamma}_{\mathrm R}
=
\operatorname{contract}
\left(
\mathcal{H}_{\mathrm R},
\mathcal{H}_{\mathrm R};
I_3,I_4
\right),
\quad
\mathcal{D}_{\mathrm R}
=
\operatorname{contract}
\left(
\mathcal{X}^{(t)},
\mathcal{H}_{\mathrm R};
I_3,I_4
\right),
\end{equation}
where
$\boldsymbol{\Gamma}_{\mathrm R}
\in
\mathbb{R}^{R_3\times R_1\times Q\times R_3\times R_1\times Q}$
and
$\mathcal{D}_{\mathrm R}
\in
\mathbb{R}^{I_1\times I_2\times R_3\times R_1\times Q}$.
The data cache is then used to construct
$\mathcal{B}_1$, $\mathcal{B}_2$, and
$\mathcal{B}_{\mathcal{C}_1}$ for updating
$\mathcal{G}_1^{(t)}$, $\mathcal{G}_2^{(t)}$, and
$\mathcal{C}_1^{(t)}$. Likewise, the Gram cache is used to construct
the corresponding Gram term tensors
$\mathcal{H}_1$, $\mathcal{H}_2$, and
$\mathcal{K}_{\mathcal{C}_1}$. The complete procedure is illustrated
in Figure~\ref{fig_2}.

Caching only the intermediate tensors $\mathcal{D}_{\mathrm R}$ and $\boldsymbol{\Gamma}_{\mathrm R}$ allows their contractions to be reused across multiple block updates, reducing redundant computation and improving update efficiency. Owing to the symmetry of the dual-core tensor wheel, the same structure-cached strategy applies to the factors in the other half. More generally, this strategy can be extended to tensor-network decompositions whose block updates share intermediate contractions.
The caching strategy extends directly to an $N$th-order tensor. Divide
the outer-ring factors into the left and right groups
$$
\left\{
\mathcal{G}_1^{(t)},\ldots,\mathcal{G}_m^{(t)}
\right\}
\quad\text{and}\quad
\left\{
\mathcal{G}_{m+1}^{(t)},\ldots,\mathcal{G}_N^{(t)}
\right\}.
$$
The left group is coupled to $\mathcal{C}_1^{(t)}$, and the right group
is coupled to $\mathcal{C}_2^{(t)}$. According to
Definition~\ref{def:dimension_set_contraction}, the contraction
dimension set for the complementary right-half subnetwork is
$$
\begin{aligned}
\Omega_{\mathcal{H}_{\mathrm R}}
={}&
\left\{
\left(
L_j^{(\mathcal{G}_j^{(t)})},
L_j^{(\mathcal{C}_2^{(t)})}
\right)
\mid j=m+1,\ldots,N
\right\}
\cup
\left\{
\left(
R_{j+1}^{(\mathcal{G}_j^{(t)})},
R_{j+1}^{(\mathcal{G}_{j+1}^{(t)})}
\right)
\mid j=m+1,\ldots,N-1
\right\}.
\end{aligned}
$$
We contract and cache this subnetwork as
$$
\mathcal{H}_{\mathrm R} 
=
\operatorname{contract}
\left(
\mathcal{C}_2^{(t)},
\mathcal{G}_{m+1}^{(t)},\ldots,\mathcal{G}_N^{(t)};
\Omega_{\mathcal{H}_{\mathrm R}}
\right),
$$
where
$
\mathcal{H}_{\mathrm R}
\in
\mathbb{R}^{
R_{m+1}\times
I_{m+1}\times\cdots\times I_N\times
R_1\times Q
}.
$
The modes $I_{m+1},\ldots,I_N$ are retained for constructing
the data and Gram caches. Let
$\widehat{\mathcal{H}}_{\mathrm R}
=\mathcal{H}_{\mathrm R}$
denote an identical copy and define
$$
\Omega_{\boldsymbol{\Gamma}_{\mathrm R}}
=
\left\{
\left(
I_j^{(\mathcal{H}_{\mathrm R})},
I_j^{(\widehat{\mathcal{H}}_{\mathrm R})}
\right)
\mid j=m+1,\ldots,N
\right\}.
$$
The Gram cache is
$$
\boldsymbol{\Gamma}_{\mathrm R}
=
\operatorname{contract}
\left(                    
\mathcal{H}_{\mathrm R},
\widehat{\mathcal{H}}_{\mathrm R};
\Omega_{\boldsymbol{\Gamma}_{\mathrm R}}
\right),
$$
where
$
\boldsymbol{\Gamma}_{\mathrm R}
\in
\mathbb{R}^{
R_{m+1}\times R_1\times Q\times
R_{m+1}\times R_1\times Q
}.
$
For the data cache, define
$$
\Omega_{\mathcal{D}_{\mathrm R}}
=
\left\{
\left(
I_j^{(\mathcal{X}^{(t)})},
I_j^{(\mathcal{H}_{\mathrm R})}
\right)
\mid j=m+1,\ldots,N
\right\}.
$$
Contracting the right-half cache with the data tensor gives
$$
\mathcal{D}_{\mathrm R}
=
\operatorname{contract}
\left(
\mathcal{X}^{(t)},
\mathcal{H}_{\mathrm R};
\Omega_{\mathcal{D}_{\mathrm R}}
\right),
$$
where
$
\mathcal{D}_{\mathrm R}
\in
\mathbb{R}^{
I_1\times\cdots\times I_m\times
R_{m+1}\times R_1\times Q
}.
$
The data cache $\mathcal{D}_{\mathrm R}$ is reused to construct
the data terms for updating
$\mathcal{G}_1^{(t)},\ldots,\mathcal{G}_m^{(t)}$ and
$\mathcal{C}_1^{(t)}$. The Gram cache
$\boldsymbol{\Gamma}_{\mathrm R}$ provides the corresponding
Gram terms. By symmetry, the left-half caches can be constructed in
the same manner for updating
$\mathcal{G}_{m+1}^{(t)},\ldots,\mathcal{G}_N^{(t)}$ and
$\mathcal{C}_2^{(t)}$.

\subsection{Computational and Memory Analysis}
This section compares explicit, implicit, and structure-cached updates in terms of computational complexity and memory consumption. The explicit, implicit, and structure-cached schemes differ in how they
construct the data terms
$\mathbf{X}_{\langle k\rangle}^{(t)}
(\mathbf{M}_{\neq k}^{(t)})^{\mathrm T}$
and
$\mathbf{x}^{(t)}
(\mathbf{N}_{\neq\mathcal{C}_1}^{(t)})^{\mathrm T}$,
as well as the corresponding Gram terms
$\mathbf{M}_{\neq k}^{(t)}
(\mathbf{M}_{\neq k}^{(t)})^{\mathrm T}$
and
$\mathbf{N}_{\neq\mathcal{C}_1}^{(t)}
(\mathbf{N}_{\neq\mathcal{C}_1}^{(t)})^{\mathrm T}$.
All remaining operations are identical, so the comparison focuses on
constructing these terms.

We first analyze the fourth-order case and then extend the results to an
$N$th-order tensor. We assume
$I_1=\cdots=I_N=I$,
$L_1=\cdots=L_N=L$, and a uniform ring rank $R$.
The reported complexity corresponds to one complete iteration and
excludes the update of $\mathcal{X}$, which is identical for all three
schemes. Computational complexity is measured in floating-point operations
(FLOPs). For
$\mathbf{A}\in\mathbb{R}^{M\times N}$ and
$\mathbf{B}\in\mathbb{R}^{N\times K}$, computing
$\mathbf{A}\mathbf{B}$ requires $MNK$ scalar multiplications and
$MK(N-1)$ scalar additions, giving an exact cost of
$MK(2N-1)$ FLOPs. After omitting the lower-order term, this cost is
approximated by $2MNK$. Hereafter, we omit the FLOP unit and use
$2MNK$ to quantify the cost of matrix multiplication.

The explicit scheme first constructs the corresponding subwheel tensor
through tensor contractions and then unfolds it to perform the matrix
operations. Consider the update of $\mathcal{G}_1^{(t)}$. The subwheel
tensor $\mathcal{M}_{\neq1}^{(t)}$ contains
$\mathcal{G}_2^{(t)}$, $\mathcal{G}_3^{(t)}$,
$\mathcal{G}_4^{(t)}$, $\mathcal{C}_1^{(t)}$, and
$\mathcal{C}_2^{(t)}$. Its construction requires four contractions
$$
\small
\begin{aligned}
\mathcal{H}_{\mathrm L}
&=
\operatorname{contract}
\left(
\mathcal{C}_1^{(t)},\mathcal{G}_2^{(t)};
L_2
\right),
\; f_1=2QL^2IR^2,
&
\mathcal{H}_{\mathrm{R},3}
&=
\operatorname{contract}
\left(
\mathcal{C}_2^{(t)},\mathcal{G}_3^{(t)};
L_3
\right),
\; f_2=2QL^2IR^2,
\\
\mathcal{H}_{\mathrm R}
&=
\operatorname{contract}
\left(
\mathcal{H}_{\mathrm{R},3},\mathcal{G}_4^{(t)};
L_4,R_4
\right),
\;f_3=2QLI^2R^3,
&
\mathcal{M}_{\neq1}^{(t)}
&=
\operatorname{contract}
\left(
\mathcal{H}_{\mathrm L},\mathcal{H}_{\mathrm R};
Q,R_3
\right),
\; f_4=2QLI^3R^3.
\end{aligned}
$$
Thus, the cost of constructing the subwheel tensor is
$$
F_{\mathcal{M}_{\neq1}}
=
4QL^2IR^2
+
2QLI^2R^3
+
2QLI^3R^3.
$$
After unfolding
$\mathcal{M}_{\neq1}^{(t)}$ into
$\mathbf{M}_{\neq1}^{(t)}
\in\mathbb{R}^{LR^2\times I^3}$,
the data term requires $2LR^2I^4$ operations, while the Gram term
requires $2L^2R^4I^3$ operations. The total cost of explicitly updating
one outer-ring factor is therefore
$$
\begin{aligned}
F_{\mathcal{G}_1}
={}&
F_{\mathcal{M}_{\neq1}}
+
2LR^2I^4
+
2L^2R^4I^3
\\
={}&
4QL^2IR^2
+
2QLI^2R^3
+
2QLI^3R^3
+
2LR^2I^4
+
2L^2R^4I^3.
\end{aligned}
$$
Under the same assumptions, the cost of explicitly updating
$\mathcal{C}_1^{(t)}$ is
$$
F_{\mathcal{C}_1}
=
2QL^2IR^2
+
2QLI^2R^3
+
2QLI^3R^3
+
2QL^2I^4R^2
+
2QL^2I^4
+
2Q^2L^4I^4.
$$
A complete iteration updates four outer-ring factors and two central
cores. Its total cost is
$$
\begin{aligned}
F_{\mathrm{explicit}}
={}&
4F_{\mathcal{G}_1}
+
2F_{\mathcal{C}_1}
=
20QL^2IR^2
+
12QLI^2R^3
+
12QLI^3R^3
+
8LR^2I^4
\\
&+
8L^2R^4I^3
+
4QL^2I^4R^2
+
4QL^2I^4
+
4Q^2L^4I^4.
\end{aligned}
$$
Because $I$ is typically much larger than $Q$, $L$, and $R$, the
terms of order $I^4$ dominate. Hence,
$$
F_{\mathrm{explicit}}
\approx
I^4
\left(
8LR^2
+
4QL^2R^2
+
4QL^2
+
4Q^2L^4
\right).
$$

The cost of implicit updating depends on the contraction order because
different orders produce intermediate tensors of different sizes. We use
the contraction order in~\eqref{eq_implicit_data_term_G1}. Under this
order, the cost of computing the data term for updating
$\mathcal{G}_1^{(t)}$ is
$$
F_{\mathrm{data}}
=
2I^4R^2L
+
2I^3R^3L^2
+
2I^2R^3L^3
+
2IR^2L^4
+
2QL^4.
$$
For the Gram term, we use the contraction order
in~\eqref{eq_implicit_gram_term_G1}. Constructing the three local Gram
tensors costs $6IR^4L^2$, while forming and connecting the two
central-core branches costs
$4QL^3R^4+4Q^2L^3R^4+4Q^2L^2R^6$. The total cost of computing the
Gram term is therefore
$$
F_{\mathrm{Gram}}
=
6IR^4L^2
+
4QL^3R^4
+
4Q^2L^3R^4
+
4Q^2L^2R^6.
$$
The total cost of implicitly updating one outer-ring factor is
$$
\begin{aligned}
F_{\mathrm{implicit}}
={}&
F_{\mathrm{data}}
+
F_{\mathrm{Gram}}
=
2I^4R^2L
+
2I^3R^3L^2
+
2I^2R^3L^3
+
2IR^2L^4
+
2QL^4
\\
&+
6IR^4L^2
+
4QL^3R^4
+
4Q^2L^3R^4
+
4Q^2L^2R^6.
\end{aligned}
$$
Similarly, the cost of implicitly updating one central core is
$$
\begin{aligned}
F_{\mathrm{implicit},\mathcal{C}_1}
={}&
2I^4R^2L
+
2I^3R^3L^2
+
2I^2R^3L^3
+
2IR^2L^4
+
2QL^4
\\
&+
8IR^4L^2
+
2QL^3R^4
+
2Q^2L^3R^4
+
4Q^2L^2R^6
+
2Q^2L^4R^4.
\end{aligned}
$$
A complete iteration updates four outer-ring factors and two central
cores. Its total cost is
$$
\begin{aligned}
F_{\mathrm{implicit}}
={}&
12I^4R^2L
+
12I^3R^3L^2
+
12I^2R^3L^3
+
12IR^2L^4
+
12QL^4
\\
&+
40IR^4L^2
+
20QL^3R^4
+
20Q^2L^3R^4
+
24Q^2L^2R^6
+
4Q^2L^4R^4.
\end{aligned}
$$
Retaining the two leading terms gives
$$
F_{\mathrm{implicit}}
\approx
12I^4R^2L
+
12I^3R^3L^2.
$$

For structure-cached updates, we first construct the right-half cache
$\mathcal{H}_{\mathrm R}$ defined in~\eqref{eq_cached_HR}. We then use
this tensor to form the data cache $\mathcal{D}_{\mathrm R}$ and the
Gram cache $\boldsymbol{\Gamma}_{\mathrm R}$ according
to~\eqref{eq_cached_gramR}. Constructing these three caches requires
$$
F_{\mathrm{cache},\mathrm R}
=
2QL^2IR^2
+2QLI^2R^3
+2QR^2I^4
+2Q^2R^4I^2.
$$
Using $\mathcal{D}_{\mathrm R}$, the data terms required to update
$\mathcal{G}_1^{(t)}$, $\mathcal{G}_2^{(t)}$, and
$\mathcal{C}_1^{(t)}$ can be constructed at a total cost of
$$
F_{\mathrm{data},\mathrm R}
=
6QLI^2R^3+6QIL^2R^2.
$$
Similarly, constructing the corresponding Gram terms from
$\boldsymbol{\Gamma}_{\mathrm R}$ requires
$$
\begin{aligned}
F_{\mathrm{Gram},\mathrm R}
={}&
8IR^4L^2
+4QL^3R^4
+4Q^2L^3R^4
+6Q^2L^2R^6
+2Q^2L^4R^4.
\end{aligned}
$$
The total cost of updating one half of the network is therefore
$$
\begin{aligned}
F_{\mathrm{cache},\mathrm{half}}
={}&
2QR^2I^4
+8QLI^2R^3
+8QL^2IR^2
+2Q^2R^4I^2\\
&+8IR^4L^2
+4QL^3R^4
+4Q^2L^3R^4
+6Q^2L^2R^6
+2Q^2L^4R^4.
\end{aligned}
$$
Because the two halves have symmetric costs, one complete
structure-cached iteration requires
$$
\begin{aligned}
F_{\mathrm{cache}}
={}&2F_{\mathrm{cache},\mathrm{half}}
=4I^4R^2Q
+16QLI^2R^3
+16QL^2IR^2
+4Q^2R^4I^2\\
&+16IR^4L^2
+8QL^3R^4
+8Q^2L^3R^4
+12Q^2L^2R^6
+4Q^2L^4R^4.
\end{aligned}
$$
Retaining the dominant terms gives
$$
F_{\mathrm{cache}}
\approx
4I^4R^2Q
+16QLI^2R^3
+4Q^2R^4I^2.
$$

The preceding fourth-order analysis extends directly to an
$N$th-order tensor. Let $N=m+s$, where
$m=\lceil N/2\rceil$ and $s=\lfloor N/2\rfloor$. The central cores
$\mathcal{C}_1$ and $\mathcal{C}_2$ are coupled to $m$ and $s$
outer-ring factors. Under the uniform settings
$I_1=\cdots=I_N=I$, $L_1=\cdots=L_N=L$, and
$R_1=\cdots=R_N=R$, the computational complexities of the three
update schemes are summarized in Table~\ref{tab:complexity}.

\begin{table}[!t]
\centering
\caption{Computational complexity comparison for an $N$th-order tensor.}
\label{tab:complexity}
\vspace{-8pt}
\setlength{\tabcolsep}{3pt}
\renewcommand{\arraystretch}{0.8}
\tiny
\resizebox{0.85\linewidth}{!}{%
\begin{tabular}{cc}
\toprule
Updating scheme & Computational complexity \\
\midrule
Explicit
&
$2I^N\!\left[
NLR^2
+Q(L^m+L^s)(R^2+1)
+Q^2(L^{2m}+L^{2s})
\right]
+\mathcal{O}(I^{N-1})$
\\

Implicit
&
$2(N+2)I^NLR^2
+2(N+2)I^{N-1}L^2R^3
+\mathcal{O}(I^{N-2})$
\\

Structure cached
&
$4QR^2I^N+\mathcal{O}\!\left(I^{\max(m,s)}\right)$
\\
\bottomrule
\end{tabular}%
}
\end{table}

\textbf{Memory complexity.}
We compare the asymptotic peak auxiliary memory required by the three
update schemes. For each scheme, peak auxiliary memory is measured by
the maximum number of tensor elements stored simultaneously during
execution and is reported using $\mathcal{O}(\cdot)$ notation. To isolate
the differences among the three implementations, we exclude the input
tensor, model parameters, and final data and Gram terms common to all
schemes. We first analyze the fourth-order case and then extend the
analysis to an $N$th-order tensor.

For explicit updates, the subring used to update an outer-ring factor
contains $LR^2I^3$ elements, while the environment tensor used to update
a central core contains $QL^2I^4$ elements. The peak auxiliary memory is
therefore
$$
S_{\mathrm{explicit}}
=
\mathcal{O}
\left(
\max
\left\{
LR^2I^3,
QL^2I^4
\right\}
\right).
$$

Implicit updates avoid constructing the complete environment tensor.
Under the adopted contraction order, the largest intermediate tensor
contains $LR^2I^3$ elements. Hence,
$$
S_{\mathrm{implicit}}
=
\mathcal{O}\left(LR^2I^3\right).
$$

For structure-cached updates, the data cache for each half contains
$QR^2I^2$ elements, and the corresponding Gram cache contains
$Q^2R^4$ elements. If the caches for both halves are retained
simultaneously, their total storage is
$$
2QR^2I^2+2Q^2R^4.
$$
The temporary half-network tensors generated during cache construction
are also of order $QR^2I^2$ and therefore do not increase the
asymptotic peak memory. Thus,
$$
S_{\mathrm{cache}}
=
\mathcal{O}
\left(
QR^2I^2+Q^2R^4
\right).
$$

When the tensor-network ranks remain fixed as $I$ increases, the
dominant memory requirements of explicit, implicit, and structure-cached
updates scale as $I^4$, $I^3$, and $I^2$. The same analysis extends to
the $N$th-order case. Let $m=\lceil N/2\rceil$ and
$s=\lfloor N/2\rfloor$, so that $m+s=N$. The resulting asymptotic peak
auxiliary memory complexities are summarized in
Table~\ref{tab:memory_complexity_N}.
%\vspace{-0.5cm}
\begin{table*}[!ht]
\centering
\caption{Peak auxiliary memory complexity comparison for
\(N\)th order tensor.}
\label{tab:memory_complexity_N}
\vspace{-8pt}
\renewcommand{\arraystretch}{1.3}
\resizebox{\textwidth}{!}{
\begin{tabular}{c c c c}
\hline
Updating scheme
& Explicit
& Implicit
& Structure cached
\\
\hline
Memory complexity
&
\(\displaystyle
\mathcal{O}\!\left(
\max\left\{
LR^2I^{N-1},
QL^mI^N,
QL^sI^N
\right\}
\right)
\)
&
\(\displaystyle
\mathcal{O}\!\left(
LR^2I^{N-1}
\right)
\)
&
\(\displaystyle
\mathcal{O}\!\left(
QR^2(I^m+I^s)+Q^2R^4
\right)
\)
\\
\hline
\end{tabular}
}
\end{table*}

\section{Numerical Experiments}\label{sect_5}

In this section, we evaluate the proposed method on real-world datasets
for tensor completion and tensor robust principal component analysis.
We also examine iterative efficiency and memory consumption to assess
the effects of the dual-core tensor wheel structure and the
structure-cached strategy. All experiments were conducted in MATLAB on
a Windows 11 desktop equipped with an AMD Ryzen 7 9800X3D processor
running at 4.7~GHz and 32~GB of RAM.

\subsection{Computational efficiency}
We evaluate the structure-cached update by comparing its runtime and
additional memory consumption with those of the explicit and implicit
updates. Each method--rank configuration is run in a separate MATLAB
process to avoid memory reuse and cached allocations. All methods use
the same observed tensor, sampling pattern, rank configuration, random
seed, and number of iterations.

\textbf{Acceleration evaluation.}
After a lightweight low-rank warm-up, we time one iteration in each
process and repeat the experiment independently ten times. The ten
measurements are summed to obtain the reported runtime for ten
iterations. The timing covers only the iterative updates and excludes
data loading, initialization, process startup, result saving, and
cooldowns. Let $T_{\mathrm{exp}}$, $T_{\mathrm{imp}}$, and
$T_{\mathrm{cache}}$ denote the runtimes of the explicit, implicit, and
structure-cached updates. The corresponding speedup factors are defined
as
$$
S_{\mathrm{exp}}
=
\frac{T_{\mathrm{exp}}}{T_{\mathrm{cache}}},
\qquad
S_{\mathrm{imp}}
=
\frac{T_{\mathrm{imp}}}{T_{\mathrm{cache}}}.
$$
A value greater than one indicates that the structure-cached update is
faster. Table~\ref{tab:acceleration} reports the measured runtimes and
speedup factors for different rank configurations across the datasets.

\begin{table}[!ht]
      \fontsize{9}{11}\selectfont
       \setlength{\tabcolsep}{2pt}
      \setlength{\abovecaptionskip}{0pt}
      \setlength{\belowcaptionskip}{0pt}
    \centering
    \caption{Runtime and speedup comparison under different rank configurations}
    \label{tab:acceleration}
    \resizebox{\textwidth}{!}{
    \begin{tabular}{lccccccc}
        \toprule
        \multicolumn{8}{c}{
            \emph{akiyo}
            $\left(144\times 176\times3\times 50\right)$
        } \\

        \midrule
        & $\mathbf{R}=[1,1,1,1]$
        & $\mathbf{R}=[2,2,2,2]$
        & $\mathbf{R}=[3,3,3,3]$
        & $\mathbf{R}=[4,4,4,4]$
        & $\mathbf{R}=[5,5,5,5]$
        & $\mathbf{R}=[6,6,6,6]$
        & $\mathbf{R}=[7,7,7,7]$ \\
        \textbf{Rank}
        & $\mathbf{L}=[1,1,1,1]$
        & $\mathbf{L}=[2,2,2,2]$
        & $\mathbf{L}=[3,3,3,3]$
        & $\mathbf{L}=[4,4,4,4]$
        & $\mathbf{L}=[5,5,5,5]$
        & $\mathbf{L}=[6,6,6,6]$
        & $\mathbf{L}=[7,7,7,7]$\\
        & $Q=1$
        & $Q=1$
        & $Q=1$
        & $Q=1$
        & $Q=1$
        & $Q=1$
        & $Q=1$ \\

        \midrule
        $T_{\mathrm{exp}}$
        & 0.492 & 1.598 & 5.768 & 21.681
        & 40.988 & 63.128 & 81.351 \\
        $T_{\mathrm{imp}}$
        & 0.804 & 1.289 & 1.513 & 1.829
        & 2.481 & 3.654 & 5.666 \\
        $T_{\mathrm{cache}}$
       & 0.550 & 0.779 & 0.815 & 0.868
        & 0.964 & 1.002 & 1.116 \\
        \midrule
       \textbf{$S_{\mathrm{exp}}$}
        & 0.895 & 2.051 & 7.080 & 24.993
        & 42.532 & 63.021 & 72.902 \\
        \textbf{$S_{\mathrm{imp}}$}
        & 1.461 & 1.654 & 1.857 & 2.109
        & 2.575 & 3.648 & 5.077 \\
        \midrule
        \addlinespace[3pt]
        \multicolumn{8}{c}{
    \emph{cars}
            $\left(256\times256\times15\times10\right)$
        } \\

        \midrule
        & $\mathbf{R}=[1,1,1,1]$
        & $\mathbf{R}=[2,2,2,2]$
        & $\mathbf{R}=[3,3,3,3]$
        & $\mathbf{R}=[4,4,4,4]$
        & $\mathbf{R}=[5,5,5,5]$
        & $\mathbf{R}=[6,6,6,6]$
        & $\mathbf{R}=[7,7,7,7]$ \\
        \textbf{Rank}
        & $\mathbf{L}=[1,1,1,1]$
        & $\mathbf{L}=[2,2,2,2]$
        & $\mathbf{L}=[3,3,3,3]$
        & $\mathbf{L}=[4,4,4,4]$
        & $\mathbf{L}=[5,5,5,5]]$
        & $\mathbf{L}=[6,6,6,6]$
        & $\mathbf{L}=[7,7,7,7]$\\
        & $Q=1$
        & $Q=1$
        & $Q=1$
        & $Q=1$
        & $Q=1$
        & $Q=1$
        & $Q=1$ \\

        \midrule
        $T_{\mathrm{exp}}$
        & 0.943 & 3.080 & 9.827 & 27.290
        & 63.119 & 95.454 & 137.170 \\
       $T_{\mathrm{imp}}$
        & 1.390 & 2.031 & 2.763 & 4.071
        & 6.503 & 10.158 & 16.519 \\
        $T_{\mathrm{cache}}$
        & 0.887 & 1.147 & 1.337 & 1.366
        & 1.501 & 1.620 & 1.837 \\
        \midrule
       \textbf{$S_{\mathrm{exp}}$}
        & 1.063 & 2.686 & 7.349 & 19.985
        & 42.046 & 58.940 & 74.691 \\
        \textbf{$S_{\mathrm{imp}}$}
        & 1.567 & 1.772 & 2.066 & 2.981
        & 4.332 & 6.272 & 8.995 \\
         \midrule
        \addlinespace[3pt]
        \multicolumn{8}{c}{
            \emph{ukulele}
            $\left(500\times500\times3\times24\right)$
        } \\
        \midrule
        & $\mathbf{R}=[1,1,1,1]$
        & $\mathbf{R}=[2,2,2,2]$
        & $\mathbf{R}=[3,3,3,3]$
        & $\mathbf{R}=[4,4,4,4]$
        & $\mathbf{R}=[5,5,5,5]$
        & $\mathbf{R}=[6,6,6,6]$
        & $\mathbf{R}=[7,7,7,7]$ \\
        \textbf{Rank}
        & $\mathbf{L}=[1,1,1,1]$
        & $\mathbf{L}=[2,2,2,2]$
        & $\mathbf{L}=[3,3,3,3]$
        & $\mathbf{L}=[4,4,4,4]$
        & $\mathbf{L}=[5,5,5,5]$
        & $\mathbf{L}=[6,6,6,6]$
        & $\mathbf{L}=[7,7,7,7]$\\
        & $Q=1$
        & $Q=1$
        & $Q=1$
        & $Q=1$
        & $Q=1$
        & $Q=1$
        & $Q=1$ \\
        \midrule
        $T_{\mathrm{exp}}$
       & 1.656 & 6.158 & 37.185 & 125.520
        & 248.210 & 303.720 & 547.860 \\
       $T_{\mathrm{imp}}$
        & 1.997 & 2.626 & 3.510 & 4.282
        & 6.075 & 9.232 & 11.815 \\
        $T_{\mathrm{cache}}$
        & 1.406 & 1.763 & 2.109 & 2.428
        & 2.950 & 3.423 & 4.287 \\
        \midrule
       \textbf{$S_{\mathrm{exp}}$}
        & 1.178 & 3.494 & 17.632 & 51.705
        & 84.128 & 88.740 & 127.784 \\
        \textbf{$S_{\mathrm{imp}}$}
        & 1.420 & 1.490 & 1.664 & 1.764
        & 2.059 & 2.697 & 2.756 \\
        \bottomrule
    \end{tabular}
    }
\end{table}

\textbf{Results analysis.}
As shown in Table~\ref{tab:acceleration}, the runtime of the explicit
update increases more rapidly with the ranks and tensor size than those
of the implicit and structure-cached updates. Except for the smallest
rank configuration on \emph{akiyo}, the structure-cached update is
faster than the explicit update in all tested settings. At $R=L=7$, its
speedup factors over the explicit update are $72.902$, $74.691$, and
$127.784$ across the three datasets. The corresponding speedups over
the implicit update are $5.077$, $8.995$, and $2.756$. These results
indicate that structure caching reduces runtime relative to both update
schemes, with larger gains at higher ranks and tensor sizes.

\textbf{Memory evaluation.}
For the memory evaluation in double precision, each method rank configuration is again run
in a separate MATLAB process. After loading the input data and
immediately before the iterative update, we record the private memory
usage as the baseline. During the update, private memory usage is
sampled periodically, and its maximum is recorded as the peak value.
Each experiment is repeated independently five times, and the mean result is reported. For each method
$q\in\{\mathrm{exp},\mathrm{imp},\mathrm{cache}\}$, the additional peak
memory consumption is defined as
$$
M_{\mathrm{add}}^{(q)}
=
\frac{
M_{\mathrm{peak}}^{(q)}-M_{\mathrm{base}}^{(q)}
}{1024^2}
\ \mathrm{MiB}.
$$
Here, $q=\mathrm{exp}$, $q=\mathrm{imp}$, and $q=\mathrm{cache}$
identify the explicit, implicit, and structure-cached updates. This
quantity measures the maximum increase in private memory during the
iterative update relative to the post-loading baseline, thereby
excluding the memory already occupied by the MATLAB process and input
data.

\begin{table}[!ht]
   \fontsize{9}{11}\selectfont
   \setlength{\tabcolsep}{2pt}
  \setlength{\abovecaptionskip}{0pt}
  \setlength{\belowcaptionskip}{0pt}
    \centering
    \caption{Additional peak memory consumption under different rank configurations}
    \label{tab:memory}
    \resizebox{\textwidth}{!}{
    \begin{tabular}{lccccccc}
        \toprule

       \multicolumn{8}{c}{
            \emph{akiyo}
            $\left(144\times176\times3\times50\right)$ \quad data memory $29.0$ MiB
            \quad \textbf{baseline memory 1185.5 MiB }
        } \\

        \midrule
        &
        $\mathbf{R}=[1,1,1,1]$ &
        $\mathbf{R}=[2,2,2,2]$ &
        $\mathbf{R}=[3,3,3,3]$ &
        $\mathbf{R}=[4,4,4,4]$ &
        $\mathbf{R}=[5,5,5,5]$ &
        $\mathbf{R}=[6,6,6,6]$ &
        $\mathbf{R}=[7,7,7,7]$ \\

        \textbf{Rank} &
        $\mathbf{L}=[1,1,1,1]$ &
        $\mathbf{L}=[2,2,2,2]$ &
        $\mathbf{L}=[3,3,3,3]$ &
        $\mathbf{L}=[4,4,4,4]$ &
        $\mathbf{L}=[5,5,5,5]$ &
        $\mathbf{L}=[6,6,6,6]$ &
        $\mathbf{L}=[7,7,7,7]$ \\

        &
        $Q=1$ & $Q=1$ & $Q=1$ & $Q=1$ &
        $Q=1$ & $Q=1$ & $Q=1$ \\

        \midrule
        % \textbf{Explicit baseline (MiB)}
        % & 1187.200 & 1180.000 & 1164.100 & 1175.200
        % & 1163.800 & 1172.100 & 1150.900 \\
       $ M_{\mathrm{add}}^{(exp)}$ 
        & 190.6& 331.8 & 705.1 & 1413.2
        & 2608.0 & 4379.2 & 5922.4 \\

       % \textbf{Implicit baseline (MiB)}
       %  & 1191.900 & 1156.300 & 1155.700 & 1152.300
       %  & 1160.800 & 1167.200 & 1161.400 \\
       $ M_{\mathrm{add}}^{(imp)}$
        & 222.4 & 182.4 & 190.7 & 187.6
        & 311.2 & 376.8 & 519.2\\

        % \textbf{Structure-caching baseline (MiB)}
        % & 1160.300 & 1150.100 & 1178.900 & 1166.300
        % & 1160.500 & 1150.400 & 1164.500 \\
        $ M_{\mathrm{add}}^{(cache)}$ 
        & 262.9 & 250.7 & 238.7 & 255.8
        & 276.2 & 230.3 & 291.1 \\
        \midrule
        \addlinespace[3pt]
         \multicolumn{8}{c}{
            \emph{cars}
            $\left(256\times256\times15\times10\right)$ \quad data memory $75.0$ MiB
            \quad \textbf{baseline memory 1314.2 MiB}
        } \\
        \midrule
        &
        $\mathbf{R}=[1,1,1,1]$ &
        $\mathbf{R}=[2,2,2,2]$ &
        $\mathbf{R}=[3,3,3,3]$ &
        $\mathbf{R}=[4,4,4,4]$ &
        $\mathbf{R}=[5,5,5,5]$ &
        $\mathbf{R}=[6,6,6,6]$ &
        $\mathbf{R}=[7,7,7,7]$ \\
        \textbf{Rank} &
        $\mathbf{L}=[1,1,1,1]$ &
        $\mathbf{L}=[2,2,2,2]$ &
        $\mathbf{L}=[3,3,3,3]$ &
        $\mathbf{L}=[4,4,4,4]$ &
        $\mathbf{L}=[5,5,5,5]$ &
        $\mathbf{L}=[6,6,6,6]$ &
        $\mathbf{L}=[7,7,7,7]$ \\
        &
        $Q=1$ & $Q=1$ & $Q=1$ & $Q=1$ &
        $Q=1$ & $Q=1$ & $Q=1$ \\
        \midrule
       % \textbf{Explicit baseline (MiB)}
       %  & 1296.300 & 1292.600 & 1296.100 & 1296.900
       %  & 1306.600 & 1290.700 & 1302.000 \\
        $ M_{\mathrm{add}}^{(exp)}$ 
        & 204.9 & 725.9 & 1643.5 & 2978.4
        & 4788.1 & 7119.2 & 10045.4 \\
       % \textbf{Implicit baseline (MiB)}
       %  & 1293.700 & 1306.900 & 1296.800 & 1302.800
       %  & 1285.700 & 1288.800 & 1295.300 \\
        $ M_{\mathrm{add}}^{(imp)}$
        & 269.0 & 236.8 & 536.1 & 1096.9
        & 2021.6 & 3397.5 & 5317.5 \\

        % \textbf{Structure-caching baseline (MiB)}
        % & 1294.900 & 1300.700 & 1290.400 & 1293.100
        % & 1292.400 & 1291.600 & 1297.800 \\
        $ M_{\mathrm{add}}^{(cache)}$  
        & 225.5 & 250.7 & 212.7 & 255.9
        & 258.3 & 279.6 & 277.1 \\

         \midrule
        \addlinespace[3pt]
         \multicolumn{8}{c}{
            \emph{ukulele}
            $\left(500\times500\times3\times24\right)$ \quad data memory $137.3$ MiB
             \quad \textbf{baseline memory 1380.0 MiB}
        } \\
        \midrule
        &
        $\mathbf{R}=[1,1,1,1]$ &
        $\mathbf{R}=[2,2,2,2]$ &
        $\mathbf{R}=[3,3,3,3]$ &
        $\mathbf{R}=[4,4,4,4]$ &
        $\mathbf{R}=[5,5,5,5]$ &
        $\mathbf{R}=[6,6,6,6]$ &
        $\mathbf{R}=[7,7,7,7]$ \\
        \textbf{Rank} &
        $\mathbf{L}=[1,1,1,1]$ &
        $\mathbf{L}=[2,2,2,2]$ &
        $\mathbf{L}=[2,2,2,2]$ &
        $\mathbf{L}=[4,4,4,4]$ &
        $\mathbf{L}=[5,5,5,5]$ &
        $\mathbf{L}=[6,6,6,6]$ &
        $\mathbf{L}=[7,7,7,7]$ \\
        &
        $Q=1$ & $Q=1$ & $Q=1$ & $Q=1$ &
        $Q=1$ & $Q=1$ & $Q=1$ \\
        \midrule
       % \textbf{Explicit baseline (MiB)}
       %  & 1368.300 & 1368.500 & 1371.100 & 1371.900
       %  & 1376.000 & 1384.600 & 1336.700 \\
        $ M_{\mathrm{add}}^{(exp)}$ 
        & 337.3 & 1186.3 & 2700.2 & 5961.5
        & 11574.5 & 19956.6 & 31661.3 \\

       % \textbf{Implicit baseline (MiB)}
       %  & 1367.000 & 1371.400 & 1376.100 & 1375.000
       %  & 1374.500 & 1383.900 & 1376.200 \\
        $ M_{\mathrm{add}}^{(imp)}$
        & 452.4& 418.1 & 501.8 & 958.9
        & 1673.2 & 2743.2& 4250.0 \\
        
        % \textbf{Structure-caching baseline (MiB)}
        % & 1368.300 & 1370.300 & 1374.400 & 1373.000
        % & 1380.600 & 1371.300 & 1376.800 \\
        $ M_{\mathrm{add}}^{(cache)}$ 
        & 368.6& 434.9 & 384.0 & 437.3
        & 523.5 & 549.5 & 595.6 \\

        \bottomrule
    \end{tabular}
    }
\end{table}
\textbf{Results analysis.}
As shown in Table~\ref{tab:memory}, the additional peak memory of the explicit update increases sharply with both the ranks and tensor size, whereas that of the structure-cached update remains comparatively stable. 
For the \emph{cars} \footnote{\url{https://data.fau.de/public/59/91/336169159/}} and \emph{ukulele} tensors, the structure-cached update has the lowest additional peak memory for $R=L\geq3$. For the \emph{akiyo} tensor, it becomes the most memory-efficient method when $R=L\geq5$. At $R=L=7$, it reduces the additional peak memory relative to the explicit update by $95.1\%$, $97.2\%$, and $98.1\%$ for \emph{akiyo}, \emph{cars}, and \emph{ukulele}, respectively. The corresponding reductions relative to the implicit update are $43.9\%$, $94.8\%$, and $86.0\%$. These results show that the memory benefit of structure caching becomes more pronounced at higher ranks and for larger tensors. Although the peak private memory of the explicit update exceeds the $32$~GiB physical RAM for the largest \emph{ukulele} configuration, the measurement remains valid because the excess memory is supported by the operating system's virtual memory.

\subsection{Tensor completion}\label{section5.1}
In the tensor completion experiments, reconstruction quality is evaluated
using the peak signal-to-noise ratio (PSNR) and structural similarity
index (SSIM) \cite{wang2004image}. For an $N$th-order tensor
($N\geq3$), MPSNR and MSSIM denote the mean PSNR and SSIM over the last
$N-2$ modes. The baselines include five representative completion
methods based on tensor networks, namely TT \cite{oseledets2011tensor},
TR \cite{zhao2016tensor}, FCTN \cite{zheng2021fully}, TW
\cite{wu2026tensor}, and SVDinsTN \cite{zheng2026structure}, together
with two classical methods, TMacTT \cite{Bengua2017Efficient} and TRLRF
\cite{yuan2019tensor}. In all fourth-order experiments, we set the ring ranks
$(R_1,\ldots,R_4)$ of the dual-core tensor wheel to a common value
$R\in\{7,\ldots,12\}$, the radial ranks $(L_1,\ldots,L_4)$ to a common
value $L\in\{1,\ldots,10\}$, and the shared rank to
$Q\in\{1,\ldots,5\}$, with $Q=1$ used unless otherwise specified.
Progressive rank exploration initializes all rank parameters to $1$ and
gradually increases them to their prescribed values. 

We evaluate five color videos, comprising three standard small-scale
videos\footnote{\url{http://trace.eas.asu.edu/yuv/}} of size
$144\times176\times3\times50$ (height $\times$ width $\times$ channels
$\times$ frames) and two recent large-scale
videos\footnote{\url{https://www.pexels.com/discover/}}. The two
large-scale videos have sizes $500\times500\times3\times24$ and
$400\times600\times3\times24$. All five videos are evaluated at sampling
rates of $5\%$, $10\%$, and $20\%$.

\textbf{Results analysis.}  
Tables~\ref{tab_5}--\ref{tab_7} show that DCTW achieves the highest
MPSNR across all 15 combinations of videos and sampling rates,
outperforming the best competing method by $0.6101$--$2.1626$~dB. It
also obtains the highest MSSIM in 14 settings. The only exception is
\emph{mother} at a sampling rate of $20\%$, where DCTW obtains $0.9637$
compared with $0.9679$ for FCTN, a difference of $0.0042$. DCTW has the shortest or second-shortest runtime in every setting and
is consistently faster than TT, TR, TRLRF, FCTN, SVDinsTN, and TW.
TMacTT is faster in most settings, whereas DCTW has a lower runtime on
\emph{train} at a sampling rate of $5\%$. These results indicate that
DCTW improves reconstruction quality while maintaining lower
computational costs than most tensor-network baselines. Figure~\ref{figure_3} provides a visual comparison of the reconstructed
frames. DCTW preserves clearer facial details with fewer visible
distortions on \emph{Akiyo} and better retains scene structures on
\emph{train}. The enlarged regions further show its ability to recover
local details from limited observations in both small- and large-scale
videos. Together, the quantitative and visual results support the
effectiveness of DCTW across different sampling rates and video scales.
\begin{table}[!ht]
  \centering
   \fontsize{9}{11}\selectfont
   \setlength{\tabcolsep}{2pt}
  \setlength{\abovecaptionskip}{0pt}
  \setlength{\belowcaptionskip}{0pt}
  \caption{Tensor completion results on five color videos at $\mathrm{SR}=5\%$.} \label{tab_5}  
  \begin{tabular}{ccccccccccc}  
    \hline  
    Video & Metric & Observed & TT& TR&TMacTT& TRLRF & FCTN  & SVDinsTN & TW& DCTW \\  
    \hline
     & MPSNR
    & 7.3326
    & 24.5413
    & 29.8665
    & 28.2789
    & 27.3730
    & 29.4842
    & \underline{32.0624}
    & 29.1171
    & \textbf{33.8535} \\  
    \emph{akiyo} & MSSIM & 0.0158
    & 0.6982
    & 0.8588
    & 0.8281
    & 0.7940
    & 0.8431
    & \underline{0.9133}
    & 0.8426
    & \textbf{0.9352}\\ 
     & TIME(s)& --
    & 440.2
    & 467.4
    & \textbf{3.7}
    & 277.7
    & 146.0
    & 207.3
    & 390.0
    & \underline{27.1}  \\ \hline\multirow{3}{*}{\textit{mother}}
    & MPSNR
    & 6.5880
    & 26.7344
    & 31.6908
    & 29.1520
    & 28.7912
    & 31.2937
    & 23.3049
    & \underline{32.2883}
    & \textbf{33.8652} \\ 
    & MSSIM
    & 0.0091
    & 0.6998
    & 0.8448
    & 0.7865
    & 0.7670
    & 0.8168
    & 0.6485
    & \underline{0.8576}
    & \textbf{0.8742} \\
    & TIME(s)
    & --
    & 443.0
    & 469.0
    & \textbf{12.7}
    & 275.1
    & 156.2
    & 162.9
    & 827.7
    & \underline{28.7} \\ 
     \hline
    \multirow{3}{*}{\textit{grandma}}
    & MPSNR
    & 11.3287
    & 24.9915
    & 31.5774
    & 31.5762
    & 27.1725
    & \underline{32.6419}
    & 23.6210
    & 31.8007
    & \textbf{34.7659} \\
    & MSSIM
    & 0.0329
    & 0.6409
    & 0.8538
    & 0.8517
    & 0.7185
    & \underline{0.8788}
    & 0.6673
    & 0.8667
    & \textbf{0.9128} \\
    
    & TIME(s)
    & --
    & 425.5
    & 452.1
    & \textbf{13.6}
    & 265.8
    & 143.3
    & 529.3
    & 457.4
    & \underline{26.3} \\
    \hline
     \multirow{3}{*}{\textit{train}}
    & MPSNR
    & 5.6285
    & 20.1939
    & 26.4434
    & 16.9551
    & 23.9835
    & \underline{30.1251}
    & 17.1222
    & 26.0670
    & \textbf{31.7167} \\
    
    & MSSIM
    & 0.0099
    & 0.5503
    & 0.7731
    & 0.3790
    & 0.6940
    & \underline{0.8729}
    & 0.4538
    & 0.7704
    & \textbf{0.8966} \\
    
    & TIME (s)
    & --
    & 2400.2
    & 2700.4
    & \underline{256.8}
    & 1370.8
    & 1276.8
    & 1006.7
    & 1977.7
    & \textbf{141.9} \\ \hline
    \multirow{3}{*}{\textit{ukulele}}
   & MPSNR
    & 4.5278
    & 23.5895
    & 29.8089
    & 25.8665
    & 27.2832
    & \underline{32.2073}
    & 20.5767
    & 29.6188
    & \textbf{33.0818} \\
    
    & MSSIM
    & 0.0079
    & 0.6741
    & 0.8118
    & 0.7229
    & 0.7588
    & \underline{0.8623}
    & 0.4317
    & 0.8095
    & \textbf{0.8768} \\
    
    & TIME (s)
    & --
    & 2681.4
    & 2880.5
    & \textbf{130.6}
    & 1464.3
    & 1440.1
    & 892.7
    & 2516.6
    & \underline{158.8} \\ \hline
   % \multirow{3}{*}{\textit{salesman}}
   %  & MPSNR
   %  & 8.3222
   %  & 21.2620
   %  & 26.3240
   %  & 16.6913
   %  & 23.7988
   %  & \underline{26.3522}
   %  & 20.1016
   %  & 26.3337
   %  & \textbf{28.1729} \\
   %  & MSSIM
   %  & 0.0147
   %  & 0.5271
   %  & 0.7865
   %  & 0.2531
   %  & 0.6726
   %  & \underline{0.8202}
   %  & 0.5261
   %  & 0.7858
   %  & \textbf{0.8619} \\    
   %  & TIME(s)
   %  & --
   %  & 445.2
   %  & 467.4
   %  & \underline{62.7}
   %  & 273.5
   %  & 361.7
   %  & 278.8
   %  & 573.2
   %  & \textbf{27.1}  \\ \hline
    % \hline
   % \multirow{3}{*}{\textit{suize}}
   %  & MPSNR
   %  & 7.2365
   %  & 25.4022
   %  & 28.3717
   %  & 25.9043
   %  & 26.2886
   %  & 26.0877
   %  & \textbf{28.9577}
   %  & 28.5060
   %  & \underline{28.6411} \\
   %  & MSSIM
   %  & 0.0101
   %  & 0.6950
   %  & 0.7586
   %  & 0.7062
   %  & 0.7114
   %  & 0.6326
   %  & \textbf{0.7790}
   %  & \underline{0.7646}
   %  & 0.7201 \\
   %  & TIME(s)
   %  & --
   %  & 445.2
   %  & 468.5
   %  & \textbf{14.9}
   %  & 272.1
   %  & 375.0
   %  & 89.5
   %  & 567.0
   %  & \underline{28.6}\\ 
   %  \hline  
  \end{tabular}
\end{table}

\begin{table}[!ht]
  \centering
   \fontsize{9}{11}\selectfont
   \setlength{\tabcolsep}{2pt}
  \setlength{\abovecaptionskip}{0pt}
  \setlength{\belowcaptionskip}{0pt}
  \caption{Tensor completion results on five color videos at $\mathrm{SR}=10\%$.} \label{tab_6}  
  \begin{tabular}{ccccccccccc}  
    \hline  
    Video & Metric & Observed & TT& TR&TMacTT& TRLRF & FCTN  & SVDinsTN & TW& DCTW \\  
    \hline
    \multirow{3}{*}{\textit{akiyo}}
    & MPSNR
    & 7.5677
    & 24.9912
    & 31.7316
    & 29.0084
    & 28.5667
    & 35.9699
    & \underline{37.2751}
    & 32.7746
    & \textbf{37.8852} \\
    
    & MSSIM
    & 0.0249
    & 0.7231
    & 0.9054
    & 0.8591
    & 0.8360
    & 0.9569
    & \underline{0.9726}
    & 0.9253
    & \textbf{0.9728} \\
    
    & TIME (s)
    & --
    & 416.4
    & 441.2
    & \textbf{1.4}
    & 265.3
    & 157.6
    & 207.7
    & 1205.8
    & \underline{29.8} \\ \hline\multirow{3}{*}{\textit{mother}}
    & MPSNR
    & 6.8230
    & 27.1845
    & 32.7483
    & 29.8739
    & 29.2408
    & \underline{36.7067}
    & 32.1112
    & 33.7096
    & \textbf{37.3448} \\
    
    & MSSIM
    & 0.0140
    & 0.7244
    & 0.8733
    & 0.8147
    & 0.7853
    & \underline{0.9266}
    & 0.8676
    & 0.8906
    & \textbf{0.9348} \\
    
    & TIME (s)
    & --
    & 439.3
    & 464.0
    & \textbf{12.9}
    & 273.0
    & 188.7
    & 63.7
    & 1141.3
    & \underline{31.6} \\ 
     \hline
    \multirow{3}{*}{\textit{grandma}}
    & MPSNR
    & 11.5640
    & 25.4552
    & 32.6945
    & 32.5761
    & 29.3795
    & 37.5345
    & \underline{37.7567}
    & 33.4916
    & \textbf{38.5752} \\
    
    & MSSIM
    & 0.0518
    & 0.6647
    & 0.8816
    & 0.8789
    & 0.7928
    & 0.9486
    & \underline{0.9540}
    & 0.9008
    & \textbf{0.9582} \\
    
    & TIME (s)
    & --
    & 423.9
    & 447.2
    & \textbf{14.0}
    & 266.5
    & 166.5
    & 522.5
    & 665.4
    & \underline{29.4} \\ \hline
     \multirow{3}{*}{\textit{train}}
    & MPSNR
    & 5.8630
    & 20.6012
    & 27.1140
    & 25.4019
    & 24.5338
    & \underline{33.7549}
    & 30.5344
    & 27.2420
    & \textbf{35.1673} \\
    
    & MSSIM
    & 0.0163
    & 0.5791
    & 0.7999
    & 0.7365
    & 0.7207
    & \underline{0.9318}
    & 0.8807
    & 0.8074
    & \textbf{0.9432} \\
    
    & TIME (s)
    & --
    & 2468.2
    & 2487.2
    & \textbf{66.0}
    & 1306.0
    & 1452.2
    & 538.4
    & 2950.8
    & \underline{152.6} \\ \hline
    \multirow{3}{*}{\textit{ukulele}}
    & MPSNR
    & 4.7624
    & 23.9194
    & 30.5082
    & 26.3561
    & 27.8593
    & \underline{35.4219}
    & 32.5932
    & 30.4142
    & \textbf{36.2214} \\
    
    & MSSIM
    & 0.0118
    & 0.6887
    & 0.8337
    & 0.7334
    & 0.7790
    & \underline{0.9296}
    & 0.8979
    & 0.8349
    & \textbf{0.9376} \\
    
    & TIME (s)
    & --
    & 2312.4
    & 2746.8
    & \textbf{14.0}
    & 1415.0
    & 1517.5
    & 531.8
    & 3374.9
    & \underline{171.3} \\\hline
   % \multirow{3}{*}{\textit{salesman}}
   %  & MPSNR
   %  & 8.5566
   %  & 21.7033
   %  & 27.4791
   %  & 22.6576
   %  & 24.9363
   %  & \underline{31.4776}
   %  & 30.1216
   %  & 27.4951
   %  & \textbf{31.6533} \\
    
   %  & MSSIM
   %  & 0.0268
   %  & 0.5632
   %  & 0.8291
   %  & 0.6178
   %  & 0.7335
   %  & \underline{0.9268}
   %  & 0.9007
   %  & 0.8299
   %  & \textbf{0.9311} \\
    
   %  & TIME (s)
   %  & --
   %  & 425.7
   %  & 449.3
   %  & \underline{45.8}
   %  & 267.0
   %  & 157.9
   %  & 170.9
   %  & 664.2
   %  & \textbf{29.1}   \\ \hline
   % \multirow{3}{*}{\textit{suize}}
   %  & MPSNR
   %  & 7.2365
   %  & 25.4022
   %  & 28.3717
   %  & 25.9043
   %  & 26.2886
   %  & 26.0877
   %  & \textbf{28.9577}
   %  & 28.5060
   %  & \underline{28.6411} \\
   %  & MSSIM
   %  & 0.0101
   %  & 0.6950
   %  & 0.7586
   %  & 0.7062
   %  & 0.7114
   %  & 0.6326
   %  & \textbf{0.7790}
   %  & \underline{0.7646}
   %  & 0.7201 \\
   %  & TIME(s)
   %  & --
   %  & 445.2
   %  & 468.5
   %  & \textbf{14.9}
   %  & 272.1
   %  & 375.0
   %  & 89.5
   %  & 567.0
   %  & \underline{28.6}\\ 
   %  \hline  
  \end{tabular}
\end{table}

\begin{table}[!ht]
  \centering
   \fontsize{9}{11}\selectfont
   \setlength{\tabcolsep}{2pt}
  \setlength{\abovecaptionskip}{0pt}
  \setlength{\belowcaptionskip}{0pt}
  \caption{Tensor completion results on five color videos at $\mathrm{SR}=20\%$.} \label{tab_7}  
  \begin{tabular}{ccccccccccc}  
    \hline  
    Video & Metric & Observed & TT& TR&TMacTT& TRLRF & FCTN  & SVDinsTN & TW& DCTW \\  
    \hline
    \multirow{3}{*}{\textit{akiyo}}
    & MPSNR
    & 8.0792
    & 25.6019
    & 32.6546
    & 31.1900
    & 29.3156
    & \underline{40.6872}
    & 31.7684
    & 34.3240
    & \textbf{42.3853} \\
    & MSSIM
    & 0.0406
    & 0.7487
    & 0.9221
    & 0.9033
    & 0.8563
    & \underline{0.9824}
    & 0.9059
    & 0.9466
    & \textbf{0.9990} \\
    & TIME (s)
    & --
    & 231.7
    & 461.9
    & \textbf{14.7}
    & 271.9
    & 196.7
    & 95.2
    & 1353.1
    & \underline{34.2} \\ \hline\multirow{3}{*}{\textit{mother}}
    & MPSNR
    & 7.3344
    & 27.8074
    & 33.6076
    & 31.0423
    & 30.5508
    & \underline{39.9699}
    & 32.4472
    & 35.0959
    & \textbf{41.0290} \\
    
    & MSSIM
    & 0.0229
    & 0.7542
    & 0.8935
    & 0.8532
    & 0.8306
    & \textbf{0.9679}
    & 0.8770
    & 0.9167
    & \underline{0.9637} \\
    
    & TIME (s)
    & --
    & 180.0
    & 463.9
    & \textbf{13.9}
    & 273.5
    & 209.1
    & 58.9
    & 1498.5
    & \underline{43.0} \\
     \hline
    \multirow{3}{*}{\textit{grandma}}
    & MPSNR
    & 12.0762
    & 26.1203
    & 33.5937
    & 33.8186
    & 30.5338
    & \underline{40.5845}
    & 35.8047
    & 34.7136
    & \textbf{41.2920} \\
    
    & MSSIM
    & 0.0853
    & 0.6980
    & 0.9010
    & 0.9087
    & 0.8310
    & \underline{0.9714}
    & 0.9393
    & 0.9217
    & \textbf{0.9753} \\
    
    & TIME (s)
    & --
    & 425.9
    & 449.7
    & \textbf{14.8}
    & 268.6
    & 184.5
    & 97.5
    & 818.5
    & \underline{33.4} \\ \hline
   \multirow{3}{*}{\textit{train}}
    & MPSNR
    & 6.3743
    & 21.1840
    & 27.9420
    & 24.8723
    & 25.2221
    & \underline{39.2173}
    & 30.0380
    & 27.9660
    & \textbf{41.3799} \\
    
    & MSSIM
    & 0.0288
    & 0.6104
    & 0.8282
    & 0.7306
    & 0.7482
    & \underline{0.9730}
    & 0.8824
    & 0.8294
    & \textbf{0.9828} \\
    
    & TIME (s)
    & --
    & 1569.5
    & 1744.5
    & \textbf{142.1}
    & 1328.9
    & 2922.5
    & 403.6
    & 3056.8
    & \underline{204.1} \\ \hline
    \multirow{3}{*}{\textit{ukulele}}
    & MPSNR
    & 5.2740
    & 24.4775
    & 31.1905
    & 27.4319
    & 28.5465
    & \underline{39.8207}
    & 33.1079
    & 31.1589
    & \textbf{40.6878} \\
    
    & MSSIM
    & 0.0182
    & 0.6981
    & 0.8535
    & 0.7644
    & 0.7979
    & \underline{0.9707}
    & 0.9057
    & 0.8532
    & \textbf{0.9729} \\
    
    & TIME (s)
    & --
    & 1607.4
    & 1850.1
    & \textbf{152.7}
    & 1416.1
    & 3369.9
    & 618.1
    & 3485.3
    & \underline{233.3} \\\hline
   % \multirow{3}{*}{\textit{salesman}}
   %  & MPSNR
   %  & 9.0668
   %  & 22.3093
   %  & 28.2777
   %  & 27.9950
   %  & 25.7261
   %  & \underline{34.4526}
   %  & 30.1228
   %  & 30.8475
   %  & \textbf{34.8098} \\
    
   %  & MSSIM
   %  & 0.0511
   %  & 0.6116
   %  & 0.8554
   %  & 0.8501
   %  & 0.7741
   %  & \underline{0.9596}
   %  & 0.9082
   %  & 0.9194
   %  & \textbf{0.9641} \\
    
   %  & TIME (s)
   %  & --
   %  & 426.0
   %  & 450.9
   %  & \textbf{3.1}
   %  & 268.1
   %  & 171.7
   %  & 109.5
   %  & 950.1
   %  & \underline{32.7} \\ \hline
   % \multirow{3}{*}{\textit{suize}}
   %  & MPSNR
   %  & 7.2365
   %  & 25.4022
   %  & 28.3717
   %  & 25.9043
   %  & 26.2886
   %  & 26.0877
   %  & \textbf{28.9577}
   %  & 28.5060
   %  & \underline{28.6411} \\
   %  & MSSIM
   %  & 0.0101
   %  & 0.6950
   %  & 0.7586
   %  & 0.7062
   %  & 0.7114
   %  & 0.6326
   %  & \textbf{0.7790}
   %  & \underline{0.7646}
   %  & 0.7201 \\
   %  & TIME(s)
   %  & --
   %  & 445.2
   %  & 468.5
   %  & \textbf{14.9}
   %  & 272.1
   %  & 375.0
   %  & 89.5
   %  & 567.0
   %  & \underline{28.6}\\ 
   %  \hline  
  \end{tabular}
\end{table}

\begin{figure}[!ht]
\centering
\setlength{\tabcolsep}{0.5pt}
\vspace{0.3cm}
\begin{tabular}{@{}ccccc@{}}
\includegraphics[width=0.20\columnwidth]{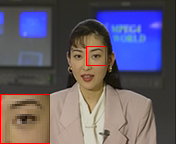} &
\includegraphics[width=0.20\columnwidth]{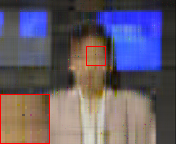} &
\includegraphics[width=0.20\columnwidth]{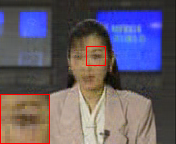} &
\includegraphics[width=0.20\columnwidth]{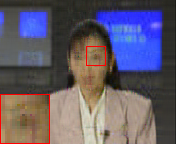} &
\includegraphics[width=0.20\columnwidth]{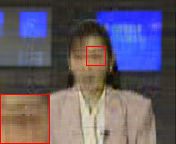}
\\[-1mm]
{\footnotesize Original image} &
{\footnotesize TT} &
{\footnotesize TR} &
{\footnotesize TMacTT} &
{\footnotesize TRLRF}
\\[1mm]
\includegraphics[width=0.20\columnwidth]{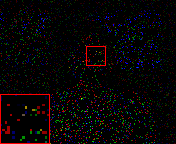} &
\includegraphics[width=0.20\columnwidth]{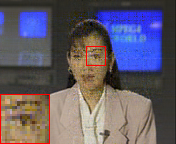} &
\includegraphics[width=0.20\columnwidth]{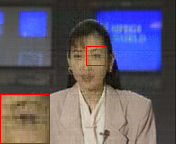} &
\includegraphics[width=0.20\columnwidth]{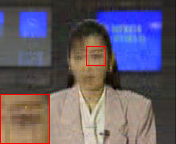} &
\includegraphics[width=0.20\columnwidth]{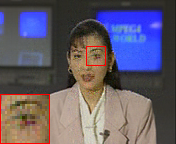}
\\[-1mm]
{\footnotesize Observed} &
{\footnotesize FCTN} &
{\footnotesize svdinsTN} &
{\footnotesize TW} &
{\footnotesize DCTW}\\[-1mm]
\includegraphics[width=0.20\columnwidth]{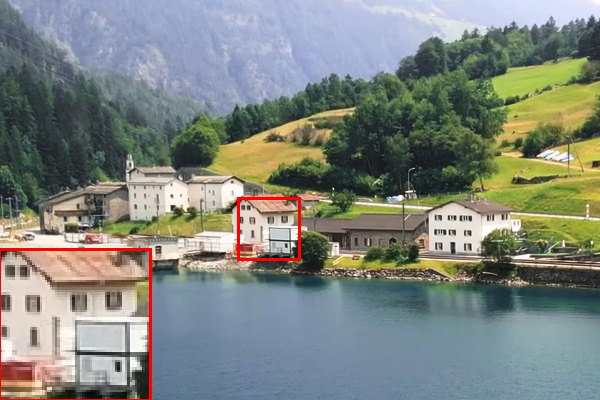} &
\includegraphics[width=0.20\columnwidth]{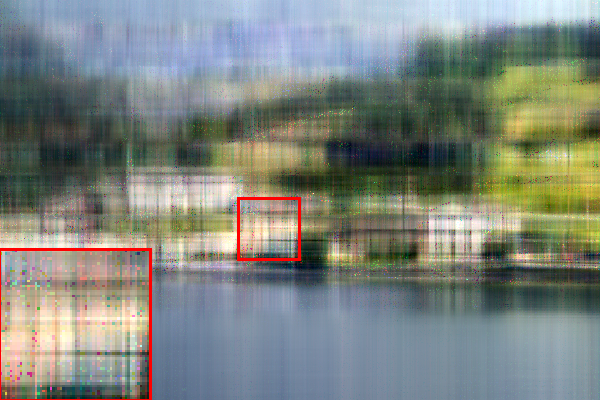} &
\includegraphics[width=0.20\columnwidth]{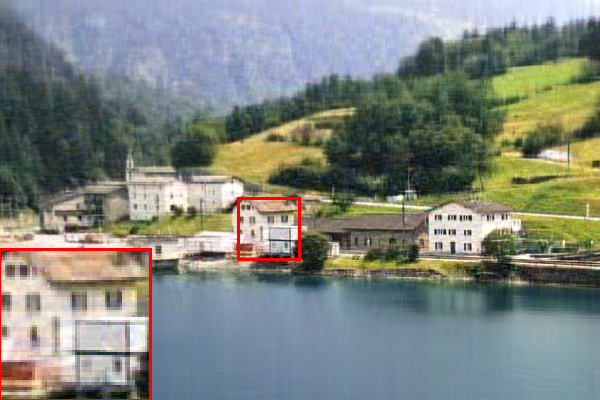} &
\includegraphics[width=0.20\columnwidth]{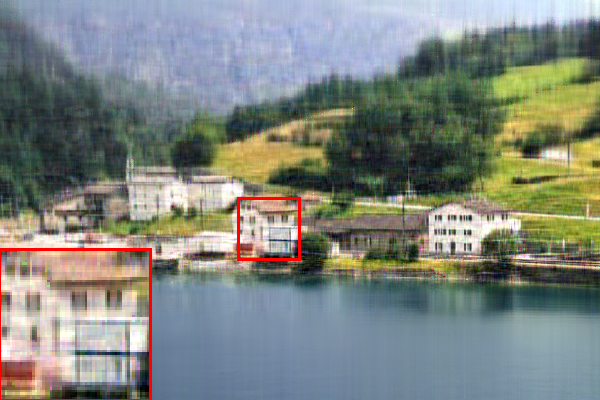} &
\includegraphics[width=0.20\columnwidth]{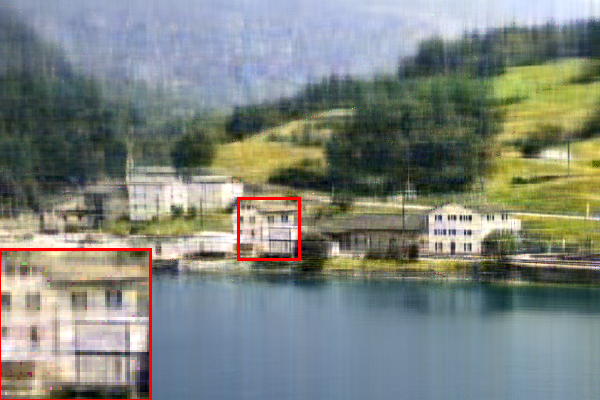}
\\
{\footnotesize Original image} &
{\footnotesize TT} &
{\footnotesize TR} &
{\footnotesize TMacTT} &
{\footnotesize TRLRF}
\\[1mm]
\includegraphics[width=0.20\columnwidth]{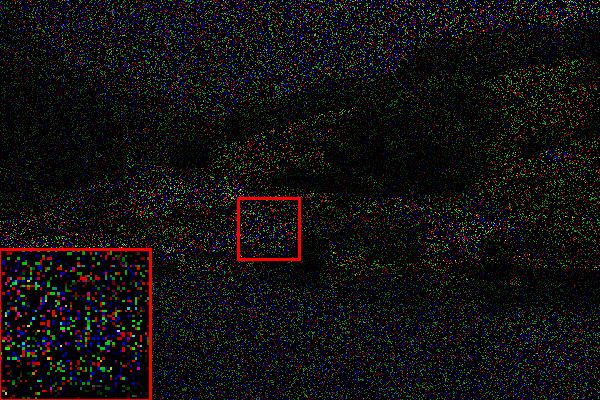} &
\includegraphics[width=0.20\columnwidth]{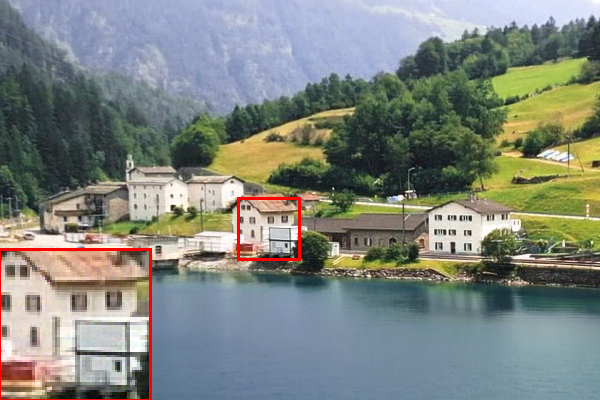} &
\includegraphics[width=0.20\columnwidth]{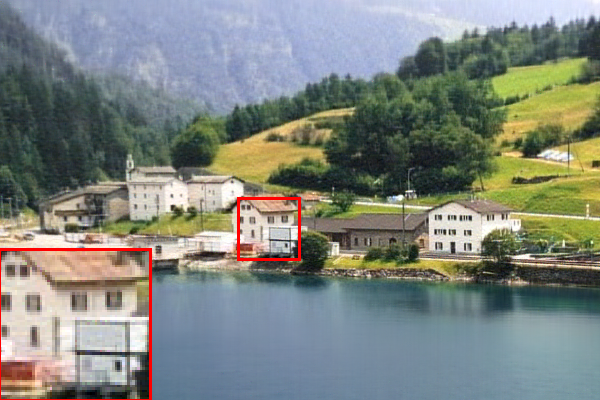} &
\includegraphics[width=0.20\columnwidth]{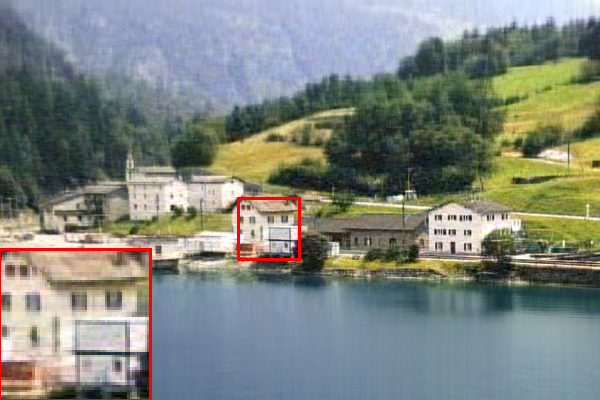}&
\includegraphics[width=0.20\columnwidth]{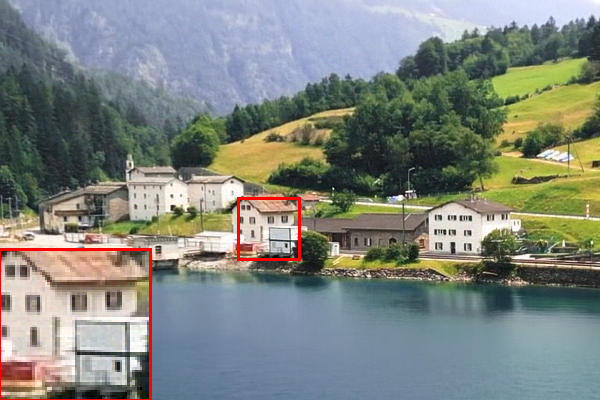}
\\[-1mm]

{\footnotesize Observed} &
{\footnotesize FCTN} &
{\footnotesize SVDinsTN} &
{\footnotesize TW} &
{\footnotesize DCTW}\\
\end{tabular}
\caption{Visualizations of the reconstructed \emph{Akiyo} and \emph{train} videos. The 25th frame of \emph{Akiyo} at a sampling rate (SR) of $5\%$ and the 12th frame of \emph{train} at an SR of $10\%$ are displayed.}\label{figure_3}
\end{figure}

\subsection{Denoising}
To assess the ability of the proposed DCTW-based TRPCA model to recover
low-rank video content from sparse corruption, we conduct denoising
experiments on four color videos: \emph{News}, \emph{Foreman},
\emph{Suzie}, and \emph{Silent}. Each video is represented as a
fourth-order tensor of size $144\times176\times3\times50$.
Salt-and-pepper noise with densities of $0.3$, $0.4$, and $0.5$ is
added randomly to each video. We compare DCTW with six representative
tensor decomposition methods: Tucker, TT, TR, SVDinsTN
\cite{zheng2026structure}, FCTN \cite{zheng2022tensor} and TW
\cite{wu2026tensor}. For Tucker, TT, TR, FCTN, and TW, we construct PAM-based denoising
variants by replacing the DCTW representation with the corresponding
tensor decomposition while retaining the same sparse-noise model,
penalty formulation, and optimization framework. This construction
allows the effect of the tensor decomposition topology to be compared
under a common denoising formulation. We use
the same rank settings and progressive adjustment strategy as those
described in Section~\ref{section5.1}.

\begin{table}[!ht]
  \centering
   \fontsize{10}{12}\selectfont
   \setlength{\tabcolsep}{2.5pt}
  \setlength{\abovecaptionskip}{0pt}
  \setlength{\belowcaptionskip}{0pt}
  \caption{Denoising performance under salt-and-pepper noise with a density of $0.3$.} \label{tab_11}  
  \begin{tabular}{cccccccccc}  
    \hline  
    Video & Metric & Observed &Tucker &TT& TR& FCTN  & SVDinsTN & TW& DCTW \\  \hline
    \multirow{3}{*}{\emph{News}}
    & MPSNR
    & 9.9987
    & 26.3918
    & 21.2012
    & 25.9363
    & 33.8353
    & \underline{35.2924}
    & 24.1166
    & \textbf{36.3619} \\
    
    & MSSIM
    & 0.1113
    & 0.8390
    & 0.6986
    & 0.8400
    & 0.9571
    & \underline{0.9596}
    & 0.7973
    & \textbf{0.9746} \\
    
    & TIME (s)
    & --
    & \textbf{70.5}
    & 732.5
    & 747.4
    & 440.0
    & 981.8
    & 535.0
    & \underline{88.3} \\ \hline
    \multirow{3}{*}{\emph{Foreman}}
    & MPSNR
    & 10.3059
    & 26.3427
    & 20.0471
    & 24.3724
    & \underline{31.3722}
    & 25.6262
    & 22.4717
    & \textbf{34.3526} \\
    
    & MSSIM
    & 0.0828
    & 0.8098
    & 0.5923
    & 0.7489
    & \underline{0.9138}
    & 0.7951
    & 0.6717
    & \textbf{0.9461} \\
    
    & TIME (s)
    & --
    & \textbf{67.1}
    & 816.1
    & 808.2
    & 515.8
    & 297.6
    & 698.5
    & \underline{97.8} \\\hline
   \multirow{3}{*}{\emph{Suzie}}
    & MPSNR
    & 10.8494
    & 31.2638
    & 25.4575
    & 29.6184
    & \underline{36.7172}
    & 32.1810
    & 30.1390
    & \textbf{39.3466} \\
    
    & MSSIM
    & 0.0471
    & 0.8704
    & 0.7545
    & 0.8428
    & \underline{0.9447}
    & 0.8925
    & 0.8589
    & \textbf{0.9638} \\
    
    & TIME (s)
    & --
    & \textbf{61.4}
    & 773.0
    & 859.6
    & 439.1
    & 365.2
    & 1387.8
    & \underline{99.8} \\ \hline   
    \multirow{3}{*}{\emph{Silent}}
    & MPSNR
    & 10.6190
    & 29.5619
    & 24.6204
    & 29.3055
    & \underline{37.8923}
    & 32.9490
    & 29.7035
    & \textbf{40.0921} \\
    
    & MSSIM
    & 0.0696
    & 0.8535
    & 0.6718
    & 0.8771
    & \underline{0.9716}
    & 0.9276
    & 0.9051
    & \textbf{0.9811} \\
    
    & TIME (s)
    & --
    & \textbf{44.4}
    & 429.3
    & 453.5
    & 345.2
    & 251.4
    & 946.2
    & \underline{73.7} \\ \hline
  \end{tabular}
\end{table}

\begin{table}[!ht]
  \centering
   \fontsize{10}{12}\selectfont
   \setlength{\tabcolsep}{2.5pt}
  \setlength{\abovecaptionskip}{0pt}
  \setlength{\belowcaptionskip}{0pt}
  \caption{Denoising performance under salt-and-pepper noise with a density of $0.4$.} \label{tab_12}  
  \begin{tabular}{cccccccccc}  
    \hline  
    Video & Metric & Observed &Tucker &TT& TR& FCTN  & SVDinsTN & TW& DCTW \\  \hline
    \multirow{3}{*}{\emph{News}}
    & MPSNR
    & 8.7501
    & 26.2492
    & 21.1556
    & 25.2943
    & 33.1376
    & \underline{34.5556}
    & 24.1312
    & \textbf{35.0654} \\
    
    & MSSIM
    & 0.0800
    & 0.8312
    & 0.6891
    & 0.8278
    & 0.9465
    & \underline{0.9468}
    & 0.7892
    & \textbf{0.9582} \\
    
    & TIME (s)
    & --
    & \textbf{54.1}
    & 764.6
    & 858.9
    & 384.4
    & 1107.7
    & 613.9
    & \underline{96.6} \\\hline
    \multirow{3}{*}{\emph{Foreman}}
    & MPSNR
    & 9.0511
    & 26.2067
    & 20.0072
    & 24.2966
    & \underline{30.6589}
    & 26.8030
    & 22.9409
    & \textbf{33.5291} \\
    
    & MSSIM
    & 0.0568
    & 0.7996
    & 0.5807
    & 0.7366
    & \underline{0.8937}
    & 0.8173
    & 0.7018
    & \textbf{0.9258} \\
    
    & TIME (s)
    & --
    & \textbf{62.9}
    & 786.9
    & 855.2
    & 427.8
    & 383.1
    & 782.9
    & \underline{94.1} \\\hline
    \multirow{3}{*}{\emph{Suize}}
    & MPSNR
    & 9.5977
    & 30.9633
    & 25.3659
    & 29.3691
    & \underline{35.5297}
    & 32.4059
    & 29.2998
    & \textbf{37.1691} \\
    
    & MSSIM
    & 0.0319
    & 0.8609
    & 0.7440
    & 0.8313
    & \underline{0.9268}
    & 0.8901
    & 0.8387
    & \textbf{0.9381} \\
    
    & TIME (s)
    & --
    & \textbf{68.6}
    & 814.3
    & 803.2
    & 510.1
    & 479.9
    & 1078.9
    & \underline{99.2} \\ \hline
    \multirow{3}{*}{\emph{Silent}}
    & MPSNR
    & 9.3687
    & 29.3590
    & 24.5323
    & 29.1939
    & \underline{36.7612}
    & 32.3170
    & 29.1804
    & \textbf{37.8125} \\
    
    & MSSIM
    & 0.0469
    & 0.8444
    & 0.6620
    & 0.8703
    & \underline{0.9620}
    & 0.9118
    & 0.8914
    & \textbf{0.9660} \\
    
    & TIME (s)
    & --
    & \textbf{80.5}
    & 687.3
    & 767.9
    & 391.3
    & 464.1
    & 996.6
    & \underline{136.3} \\ \hline
  \end{tabular}
\end{table}

\begin{table}[!ht]
  \centering
   \fontsize{10}{12}\selectfont
   \setlength{\tabcolsep}{2.5pt}
  \setlength{\abovecaptionskip}{0pt}
  \setlength{\belowcaptionskip}{0pt}
  \caption{Denoising performance under salt-and-pepper noise with a density of $0.5$.} \label{tab_13}  
  \begin{tabular}{cccccccccc}  
    \hline  
    Video & Metric & Observed &Tucker &TT& TR& FCTN  & SVDinsTN & TW& DCTW \\  \hline
    \multirow{3}{*}{\emph{News}}
    & MPSNR
    & 7.7844
    & 26.0502
    & 20.9874
    & 25.4719
    & 31.9886
    & \textbf{32.5575}
    & 23.9321
    & \underline{32.4312} \\
    
    & MSSIM
    & 0.0585
    & 0.8212
    & 0.6768
    & 0.8250
    & \textbf{0.9277}
    & \underline{0.9151}
    & 0.7811
    & 0.9122 \\
    
    & TIME (s)
    & --
    & \textbf{70.5}
    & 814.1
    & 771.1
    & 477.9
    & 845.4
    & 579.4
    & \underline{97.8} \\\hline
    \multirow{3}{*}{\emph{Foreman}}
    & MPSNR
    & 8.0830
    & 26.0636
    & 19.9730
    & 24.0151
    & \underline{29.4228}
    & 26.1093
    & 23.6114
    & \textbf{30.8232} \\
    
    & MSSIM
    & 0.0402
    & 0.7883
    & 0.5686
    & 0.7191
    & \textbf{0.8571}
    & 0.7890
    & 0.7204
    & \underline{0.8565} \\
    
    & TIME (s)
    & --
    & \textbf{70.2}
    & 723.2
    & 764.5
    & 406.4
    & 368.3
    & 956.0
    & \underline{120.9} \\\hline
   \multirow{3}{*}{\emph{Suize}}
    & MPSNR
    & 8.6283
    & 30.7261
    & 24.6700
    & 29.1648
    & \underline{33.9292}
    & 32.1404
    & 29.3530
    & \textbf{34.5807} \\
    
    & MSSIM
    & 0.0226
    & 0.8496
    & 0.7325
    & 0.8195
    & \underline{0.8975}
    & 0.8784
    & 0.8301
    & \textbf{0.9093} \\
    
    & TIME (s)
    & --
    & \textbf{43.2}
    & 440.4
    & 465.2
    & 389.8
    & 286.9
    & 713.3
    & \underline{47.9} \\ \hline 
    \multirow{3}{*}{\emph{Silent}}
    & MPSNR
    & 8.4000
    & 29.1701
    & 24.4457
    & 28.7478
    & \underline{35.1889}
    & 30.9389
    & 29.1928
    & \textbf{35.5812} \\
    
    & MSSIM
    & 0.0327
    & 0.8348
    & 0.6518
    & 0.8610
    & \underline{0.9449}
    & 0.8780
    & 0.8884
    & \textbf{0.9546} \\
    
    & TIME (s)
    & --
    & \underline{70.9}
    & 732.7
    & 690.4
    & 505.7
    & 564.1
    & 1097.8
    & \textbf{45.6} \\\hline 
  \end{tabular}
\end{table}

\begin{figure}[!ht]
\centering
\setlength{\tabcolsep}{0.5pt}
\vspace{0.3cm}
\begin{tabular}{@{}cccc@{}}
\includegraphics[width=0.23\columnwidth]{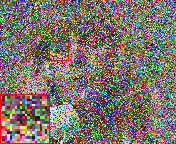} &
\includegraphics[width=0.23\columnwidth]{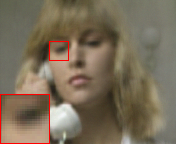} &
\includegraphics[width=0.23\columnwidth]{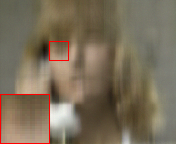} &
\includegraphics[width=0.23\columnwidth]{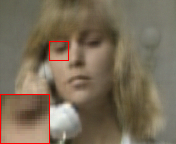}
\\[-1mm]
{\footnotesize Observed} &
{\footnotesize Tucker} &
{\footnotesize TT} &
{\footnotesize TR}
\\[-1mm]
\includegraphics[width=0.23\columnwidth]{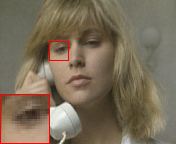} &
\includegraphics[width=0.23\columnwidth]{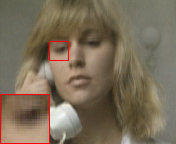} &
\includegraphics[width=0.23\columnwidth]{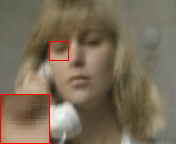} &
\includegraphics[width=0.23\columnwidth]{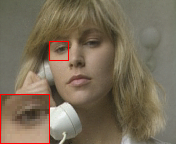} 
\\[-1mm]
{\footnotesize FCTN} &
{\footnotesize SVDinsTN} &
{\footnotesize TW} &
{\footnotesize DCTW}\\[-1mm]
\includegraphics[width=0.23\columnwidth]{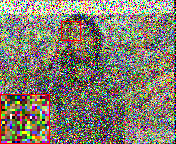} &
\includegraphics[width=0.23\columnwidth]{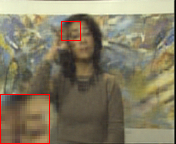} &
\includegraphics[width=0.23\columnwidth]{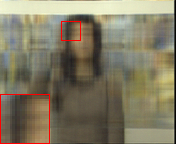} &
\includegraphics[width=0.23\columnwidth]{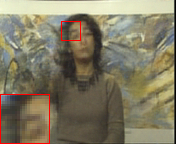}
\\
{\footnotesize Observed} &
{\footnotesize Tucker} &
{\footnotesize TT} &
{\footnotesize TR}
\\[-1mm]
\includegraphics[width=0.23\columnwidth]{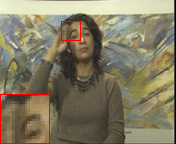} &
\includegraphics[width=0.23\columnwidth]{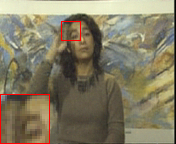} &
\includegraphics[width=0.23\columnwidth]{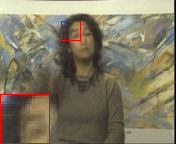} &
\includegraphics[width=0.23\columnwidth]{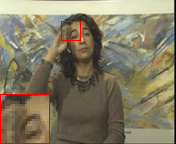}
\\[-1mm]
{\footnotesize FCTN} &
{\footnotesize SVDinsTN} &
{\footnotesize TW} &
{\footnotesize DCTW}\\
\end{tabular}
\caption{Visual comparison of the 25th frames reconstructed from the \emph{Suzie} and \emph{Silent} videos under salt-and-pepper noise with a density of $0.4$.}\label{figure_4}
\end{figure}

\textbf{Results analysis.} 
As shown in Tables~\ref{tab_11}--\ref{tab_13}, DCTW achieves the highest
MPSNR and MSSIM for all four videos at noise densities of $0.3$ and
$0.4$. Its MPSNR improvements over the best competing method range from
$0.5098$ to $2.9804$~dB. At a noise density of $0.5$, DCTW obtains the
highest MPSNR on \emph{Foreman}, \emph{Suzie}, and \emph{Silent}, and
the second-highest MPSNR on \emph{News}. It also achieves the highest
MSSIM on \emph{Suzie} and \emph{Silent}, whereas FCTN performs best on
\emph{News} and \emph{Foreman}. DCTW records the shortest or second-shortest runtime in every setting. It is consistently faster than TT, TR, FCTN, SVDinsTN, and TW, and is
second only to Tucker in most cases. On \emph{Silent} at a noise density
of $0.5$, DCTW is the fastest method. Figure~\ref{figure_4} further shows
that DCTW leaves less residual salt-and-pepper noise and better preserves
facial contours, eye details, and hair textures. Across the tested noise levels, DCTW maintains consistently high denoising accuracy while requiring less runtime than most competing methods.

\subsection{Structural topology and $Q$-rank analysis}
In this section, we compare the tensor completion performance of the
dual-core tensor wheel  and the tensor wheel under identical ring-rank and radial-rank settings to analyze how their
central core structures differ in terms of model recovery capability. We
further investigate the effect of different central-rank $Q$ values on the
completion performance of the dual-core tensor-wheel model, providing
empirical guidance for selecting $Q$. Throughout this section, PSNR refers to the mean peak signal-to-noise ratio (MPSNR).

\textbf{Structural comparison}. At a sampling rate of 10\%, we compare different tensor wheel structures
and intermediate term update strategies on the \emph{container},
\emph{mian}\footnote{\url{https://openremotesensing.net/knowledgebase/hyperspectral-video/}},
and \emph{eagle} datasets. The results are summarized in
Table~\ref{tab:structural}. Cases that cannot be executed due to insufficient
RAM memory are denoted by ``--''. All algorithms use the same ring and radial ranks
and are run for 500 iterations. The tensor wheel and the dual-core tensor wheel are compared to analyze the influence of the
central-core structure on tensor completion performance. Under the dual-core
structure, explicit construction, implicit contraction, and
structure-cached updates are further compared to evaluate the acceleration
achieved by structure-cached.

\begin{table}[!ht]
    \centering
    \setlength{\abovecaptionskip}{0pt}
    \setlength{\belowcaptionskip}{2pt}

   \caption{Comparison of tensor wheel structures and update strategies}
    \label{tab:structural}

    \fontsize{9}{11}\selectfont
    \setlength{\tabcolsep}{1.2pt}
    \renewcommand{\arraystretch}{0.92}

    \begin{adjustbox}{max width=\textwidth}
    \begin{tabular}{@{}l*{6}{cc}@{}}
        \toprule

        % ==================== Dataset 1 ====================
        \multicolumn{13}{c}{
            \emph{container}
            $\left(144\times176\times3\times50\right)$
        }\\
        \midrule

        \multirow{4}{*}{\textbf{Rank}}
        & \multicolumn{2}{c}{$\mathbf{R}=[2,2,2,2]$}
        & \multicolumn{2}{c}{$\mathbf{R}=[3,3,3,3]$}
        & \multicolumn{2}{c}{$\mathbf{R}=[4,4,4,4]$}
        & \multicolumn{2}{c}{$\mathbf{R}=[5,5,5,5]$}
        & \multicolumn{2}{c}{$\mathbf{R}=[6,6,6,6]$}
        & \multicolumn{2}{c}{$\mathbf{R}=[7,7,7,7]$}\\

        & \multicolumn{2}{c}{$\mathbf{L}=[2,2,2,2]$}
        & \multicolumn{2}{c}{$\mathbf{L}=[3,3,3,3]$}
        & \multicolumn{2}{c}{$\mathbf{L}=[3,3,3,3]$}
        & \multicolumn{2}{c}{$\mathbf{L}=[3,3,3,3]$}
        & \multicolumn{2}{c}{$\mathbf{L}=[3,3,3,3]$}
        & \multicolumn{2}{c}{$\mathbf{L}=[3,3,3,3]$}\\

        & \multicolumn{2}{c}{$Q=1$}
        & \multicolumn{2}{c}{$Q=1$}
        & \multicolumn{2}{c}{$Q=1$}
        & \multicolumn{2}{c}{$Q=1$}
        & \multicolumn{2}{c}{$Q=1$}
        & \multicolumn{2}{c}{$Q=1$}\\

        \cmidrule(lr){2-3}
        \cmidrule(lr){4-5}
        \cmidrule(lr){6-7}
        \cmidrule(lr){8-9}
        \cmidrule(lr){10-11}
        \cmidrule(l){12-13}

        & PSNR & Time (s)
        & PSNR & Time (s)
        & PSNR & Time (s)
        & PSNR & Time (s)
        & PSNR & Time (s)
        & PSNR & Time (s)\\
        \midrule

        TW
        & 23.3348 & 149.4
        & 27.0711 & 408.2
        & 26.9978 & 277.8
        & 31.2693 & 1237.2
        & 32.8254 & 1434.5
        & -- & --\\

        DCTW-exp
        & 23.2600 & 61.4 
        & 27.3923 & 230.6
        & 29.5029 & 363.8 
        & 31.5082 & 624.5
        & 33.4262 & 852.9
        & 34.9061 & 994.2 \\

        DCTW-imp
        & 23.2600 & 35.5 
        & 27.3923 & 42.1
        & 29.5029 & 47.6 
        & 31.5082 & 55.6
        & 33.4262 & 63.2
        & 34.9061 & 75.9\\

        DCTW-cache
        & 23.2600 & 19.1
        & 27.3923 & 20.3
        & 29.5029 & 21.2
        & 31.5082 & 22.7
        & 33.4262 & 24.3
        & 34.9061 & 25.8 \\

        % ==================== Dataset 2 ====================
        \midrule
        \multicolumn{13}{c}{
            \emph{mian}
            $\left(256\times256\times15\times10\right)$
        }\\
        \midrule

        \multirow{4}{*}{\textbf{Rank}}
        & \multicolumn{2}{c}{$\mathbf{R}=[2,2,2,2]$}
        & \multicolumn{2}{c}{$\mathbf{R}=[3,3,3,3]$}
        & \multicolumn{2}{c}{$\mathbf{R}=[4,4,4,4]$}
        & \multicolumn{2}{c}{$\mathbf{R}=[5,5,5,5]$}
        & \multicolumn{2}{c}{$\mathbf{R}=[6,6,6,6]$}
        & \multicolumn{2}{c}{$\mathbf{R}=[7,7,7,7]$}\\

        & \multicolumn{2}{c}{$\mathbf{L}=[2,2,2,2]$}
        & \multicolumn{2}{c}{$\mathbf{L}=[3,3,3,3]$}
        & \multicolumn{2}{c}{$\mathbf{L}=[3,3,3,3]$}
        & \multicolumn{2}{c}{$\mathbf{L}=[3,3,3,3]$}
        & \multicolumn{2}{c}{$\mathbf{L}=[3,3,3,3]$}
        & \multicolumn{2}{c}{$\mathbf{L}=[3,3,3,3]$}\\

        & \multicolumn{2}{c}{$Q=1$}
        & \multicolumn{2}{c}{$Q=1$}
        & \multicolumn{2}{c}{$Q=1$}
        & \multicolumn{2}{c}{$Q=1$}
        & \multicolumn{2}{c}{$Q=1$}
        & \multicolumn{2}{c}{$Q=1$}\\

        \cmidrule(lr){2-3}
        \cmidrule(lr){4-5}
        \cmidrule(lr){6-7}
        \cmidrule(lr){8-9}
        \cmidrule(lr){10-11}
        \cmidrule(l){12-13}

        & PSNR & Time (s)
        & PSNR & Time (s)
        & PSNR & Time (s)
        & PSNR & Time (s)
        & PSNR & Time (s)
        & PSNR & Time (s)\\
        \midrule

        TW
       & 25.1038 & 318.4
        & 25.1037 & 313.0
        & 28.3502 & 422.7
        & 35.7883 & 1269.6 
        & 37.3986 & 7508.0 
        & -- & -- \\

        DCTW-exp
        & 24.7597 & 147.7 
        & 30.4973 & 370.9
        & 33.2468 & 524.7
        &35.8126 & 844.4
        & 37.2430 & 1119.5
        & 37.9881 & 1268.4\\

        DCTW-imp
        & 24.7597 & 85.5
        & 30.4973 & 99.1
        & 33.2468 & 122.4 
        & 35.8126 & 153.3
       & 37.2430 & 191.1 
       & 37.9881 & 227.5  \\

        DCTW-cache
        & 24.7597 & 42.4
        & 30.4973 &43.8
        & 33.2468 & 45.7
        & 35.8126& 50.51
        & 37.2430 &53.9
        & 37.9881 &57.1 \\

        % ==================== Dataset 3 ====================
        \midrule
        \multicolumn{13}{c}{
            \emph{eagle}
            $\left(470\times470\times3\times25\right)$
        }\\
        \midrule

        \multirow{4}{*}{\textbf{Rank}}
        & \multicolumn{2}{c}{$\mathbf{R}=[2,2,2,2]$}
        & \multicolumn{2}{c}{$\mathbf{R}=[3,3,3,3]$}
        & \multicolumn{2}{c}{$\mathbf{R}=[4,4,4,4]$}
        & \multicolumn{2}{c}{$\mathbf{R}=[5,5,5,5]$}
        & \multicolumn{2}{c}{$\mathbf{R}=[6,6,6,6]$}
        & \multicolumn{2}{c}{$\mathbf{R}=[7,7,7,7]$}\\

        & \multicolumn{2}{c}{$\mathbf{L}=[2,2,2,2]$}
        & \multicolumn{2}{c}{$\mathbf{L}=[2,2,2,2]$}
        & \multicolumn{2}{c}{$\mathbf{L}=[2,2,2,2]$}
        & \multicolumn{2}{c}{$\mathbf{L}=[2,2,2,2]$}
        & \multicolumn{2}{c}{$\mathbf{L}=[2,2,2,2]$}
        & \multicolumn{2}{c}{$\mathbf{L}=[2,2,2,2]$}\\

        & \multicolumn{2}{c}{$Q=1$}
        & \multicolumn{2}{c}{$Q=1$}
        & \multicolumn{2}{c}{$Q=1$}
        & \multicolumn{2}{c}{$Q=1$}
        & \multicolumn{2}{c}{$Q=1$}
        & \multicolumn{2}{c}{$Q=1$}\\

        \cmidrule(lr){2-3}
        \cmidrule(lr){4-5}
        \cmidrule(lr){6-7}
        \cmidrule(lr){8-9}
        \cmidrule(lr){10-11}
        \cmidrule(l){12-13}

        & PSNR & Time (s)
        & PSNR & Time (s)
        & PSNR & Time (s)
        & PSNR & Time (s)
        & PSNR & Time (s)
        & PSNR & Time (s)\\
        \midrule

        TW
        & 23.2715 & 652.2
        & 25.5271 & 852.8
        & 25.6932 & 1210.7
        & 28.9649 & 2438.9
       & 30.3536 & 14019.7 
        & -- & -- \\

        DCTW-exp
        & 23.1424 & 272.2
        & 25.6670 & 544.1 
       & 27.5121 & 1000.5
       & 29.1493 & 1790.2 
        & 30.4482 & 2997.6
        & 31.4183 & 4013.6 \\

        DCTW-imp
        & 23.1424 & 109.2
        & 25.6670 & 113.7 
      & 27.5121 & 130.5
       & 29.1493 & 152.5  
        & 30.4482 & 164.0 
        & 31.4183 & 182.5\\ 

        DCTW-cache
        & 23.1424 & 72.8
       & 25.6670 & 81.1 
        & 27.5121 & 99.2
        & 29.1493 & 113.1 
        & 30.4482 & 120.9
        & 31.4183 & 135.5\\

        \bottomrule
    \end{tabular}
    \end{adjustbox}
\end{table}
\textbf{Results analysis.} The three DCTW update strategies compared in Table~\ref{tab:structural} yield identical PSNR values under each rank setting, confirming their numerical equivalence despite differences in computational implementation. Compared with TW, the dual-core structure generally achieves higher reconstruction quality at moderate and high ranks. Although its completion performance is slightly lower than that of TW under a few rank settings, the differences are small. Structure-cached substantially reduces the running time, achieving speedups of $3.2$--$38.5$, $3.5$--$22.2$, and $3.7$--$29.6$ over explicit  construction on \emph{container}, \emph{mian}, and \emph{eagle}, respectively, while remaining approximately $1.3$--$4.0$ faster than implicit contraction. Moreover, DCTW remains executable at $R=7$, where TW runs out of memory. 

\textbf{Influence of the \(Q\)-rank}. To investigate the effect of the \(Q\)-rank, we conduct experiments with 500 iterations on the same three datasets under a sampling rate of 10\%. The ring rank and radial rank are fixed across all configurations, while only the \(Q\)-rank is varied. Cases that fail to run due to insufficient RAM are marked with ``--''.

\begin{table}[!ht]
    \centering
    \setlength{\abovecaptionskip}{0pt}
    \setlength{\belowcaptionskip}{2pt}

   \caption{Influence of the \(Q\)-rank}
    \label{tab:Q-rank}

    \fontsize{9}{11}\selectfont
    \setlength{\tabcolsep}{1.2pt}
    \renewcommand{\arraystretch}{0.92}

    \begin{adjustbox}{max width=\textwidth}
    \begin{tabular}{@{}l*{6}{cc}@{}}
        \toprule

        % ==================== Dataset 1 ====================
        \multicolumn{13}{c}{
            \emph{container}
            $\left(144\times176\times3\times50\right)$
        }\\
        \midrule

        \multirow{4}{*}{\textbf{Rank}}
        & \multicolumn{2}{c}{$\mathbf{R}=[7,7,7,7]$}
        & \multicolumn{2}{c}{$\mathbf{R}=[7,7,7,7]$}
        & \multicolumn{2}{c}{$\mathbf{R}=[7,7,7,7]$}
        & \multicolumn{2}{c}{$\mathbf{R}=[7,7,7,7]$}
        & \multicolumn{2}{c}{$\mathbf{R}=[7,7,7,7]$}
        & \multicolumn{2}{c}{$\mathbf{R}=[7,7,7,7]$}\\

        & \multicolumn{2}{c}{$\mathbf{L}=[5,5,5,5]$}
        & \multicolumn{2}{c}{$\mathbf{L}=[5,5,5,5]$}
        & \multicolumn{2}{c}{$\mathbf{L}=[5,5,5,5]$}
        & \multicolumn{2}{c}{$\mathbf{L}=[7,7,7,7]$}
        & \multicolumn{2}{c}{$\mathbf{L}=[7,7,7,7]$}
        & \multicolumn{2}{c}{$\mathbf{L}=[7,7,7,7]$}\\

        & \multicolumn{2}{c}{$Q=1$}
        & \multicolumn{2}{c}{$Q=2$}
        & \multicolumn{2}{c}{$Q=3$}
        & \multicolumn{2}{c}{$Q=1$}
        & \multicolumn{2}{c}{$Q=2$}
        & \multicolumn{2}{c}{$Q=3$}\\

        \cmidrule(lr){2-3}
        \cmidrule(lr){4-5}
        \cmidrule(lr){6-7}
        \cmidrule(lr){8-9}
        \cmidrule(lr){10-11}
        \cmidrule(l){12-13}

        & PSNR & Time (s)
        & PSNR & Time (s)
        & PSNR & Time (s)
        & PSNR & Time (s)
        & PSNR & Time (s)
        & PSNR & Time (s)\\
        \midrule

       DCTW-exp
       & 37.2224 & 1625.1 
        & 36.0153  &  2308.7
        & 36.4294 & 2937.4
        & 38.3486 & 2269.3 
        & 37.3358 & 4107.4
        & 37.1021 & 6973.4 \\

        DCTW-imp
        & 37.2224 & 115.2  
        & 36.0153 & 142.1  
        & 36.4294 & 166.9 
        & 38.3486 & 162.3 
        &37.3358 & 203.9
        & 37.1021 & 394.3 \\ 

        DCTW-cache
        & 37.2224 & 26.6   
       & 36.0153 & 35.5
        & 36.4294 & 44.1 
        &38.3486 & 28.4  
        & 37.3358 & 38.0
        & 37.1021 & 79.3 \\

        % ==================== Dataset 2 ====================
        \midrule
        \multicolumn{13}{c}{
            \emph{mian}
            $\left(256\times256\times15\times10\right)$
        }\\
        \midrule

        \multirow{4}{*}{\textbf{Rank}}
      & \multicolumn{2}{c}{$\mathbf{R}=[7,7,7,7]$}
        & \multicolumn{2}{c}{$\mathbf{R}=[7,7,7,7]$}
        & \multicolumn{2}{c}{$\mathbf{R}=[7,7,7,7]$}
        & \multicolumn{2}{c}{$\mathbf{R}=[7,7,7,7]$}
        & \multicolumn{2}{c}{$\mathbf{R}=[7,7,7,7]$}
        & \multicolumn{2}{c}{$\mathbf{R}=[7,7,7,7]$}\\

        & \multicolumn{2}{c}{$\mathbf{L}=[5,5,5,5]$}
        & \multicolumn{2}{c}{$\mathbf{L}=[5,5,5,5]$}
        & \multicolumn{2}{c}{$\mathbf{L}=[5,5,5,5]$}
        & \multicolumn{2}{c}{$\mathbf{L}=[7,7,7,7]$}
        & \multicolumn{2}{c}{$\mathbf{L}=[7,7,7,7]$}
        & \multicolumn{2}{c}{$\mathbf{L}=[7,7,7,7]$}\\

        & \multicolumn{2}{c}{$Q=1$}
        & \multicolumn{2}{c}{$Q=2$}
        & \multicolumn{2}{c}{$Q=3$}
        & \multicolumn{2}{c}{$Q=1$}
        & \multicolumn{2}{c}{$Q=2$}
        & \multicolumn{2}{c}{$Q=3$}\\

        \cmidrule(lr){2-3}
        \cmidrule(lr){4-5}
        \cmidrule(lr){6-7}
        \cmidrule(lr){8-9}
        \cmidrule(lr){10-11}
        \cmidrule(l){12-13}

        & PSNR & Time (s)
        & PSNR & Time (s)
        & PSNR & Time (s)
        & PSNR & Time (s)
        & PSNR & Time (s)
        & PSNR & Time (s)\\
        \midrule

        DCTW-exp
        & 38.9507 & 2561.4
        & 39.1955 & 3896.6
        & 39.5448 & 5847.8 
        & 39.3417 & 3838.9 
        & 39.5736 & 7987.9 
        & 40.0027 & 8720.2   \\

        DCTW-imp
       & 38.9507 & 347.6 
        & 39.1955 & 367.4  
        & 39.5448 & 378.2 
        & 39.3417 & 483.3 
        & 39.5736 & 519.5 
        & 40.0027 & 519.8    \\

        DCTW-cache
        & 38.9507 & 61.3 
        & 39.1955 & 78.9 
        & 39.5448 & 95.5 
        & 39.3417 & 61.6 
        & 39.5736 & 82.5 
        & 40.0027 & 98.0  \\

        % ==================== Dataset 3 ====================
        \midrule
        \multicolumn{13}{c}{
            \emph{eagle}
            $\left(470\times470\times3\times25\right)$
        }\\
        \midrule

     \multirow{4}{*}{\textbf{Rank}}
        & \multicolumn{2}{c}{$\mathbf{R}=[7,7,7,7]$}
        & \multicolumn{2}{c}{$\mathbf{R}=[7,7,7,7]$}
        & \multicolumn{2}{c}{$\mathbf{R}=[7,7,7,7]$}
        & \multicolumn{2}{c}{$\mathbf{R}=[7,7,7,7]$}
        & \multicolumn{2}{c}{$\mathbf{R}=[7,7,7,7]$}
        & \multicolumn{2}{c}{$\mathbf{R}=[7,7,7,7]$}\\

        & \multicolumn{2}{c}{$\mathbf{L}=[5,5,5,5]$}
        & \multicolumn{2}{c}{$\mathbf{L}=[5,5,5,5]$}
        & \multicolumn{2}{c}{$\mathbf{L}=[5,5,5,5]$}
        & \multicolumn{2}{c}{$\mathbf{L}=[7,7,7,7]$}
        & \multicolumn{2}{c}{$\mathbf{L}=[7,7,7,7]$}
        & \multicolumn{2}{c}{$\mathbf{L}=[7,7,7,7]$}\\

        & \multicolumn{2}{c}{$Q=1$}
        & \multicolumn{2}{c}{$Q=2$}
        & \multicolumn{2}{c}{$Q=3$}
        & \multicolumn{2}{c}{$Q=1$}
        & \multicolumn{2}{c}{$Q=2$}
        & \multicolumn{2}{c}{$Q=3$}\\

        \cmidrule(lr){2-3}
        \cmidrule(lr){4-5}
        \cmidrule(lr){6-7}
        \cmidrule(lr){8-9}
        \cmidrule(lr){10-11}
        \cmidrule(l){12-13}

        & PSNR & Time (s)
        & PSNR & Time (s)
        & PSNR & Time (s)
        & PSNR & Time (s)
        & PSNR & Time (s)
        & PSNR & Time (s)\\
        \midrule

        DCTW-exp
        & 34.2132 & 7795.2  
        & 33.8163 & 9897.9
        & -- & -- 
        & 34.7246 & 28378.0 
        & -- & -- 
        & -- & -- \\

        DCTW-imp
        & 34.2132 & 318.2 
        & 33.8163 & 361.1
        & 34.1025 & 406.2 
        & 34.7246 & 410.2
        & 34.3188 & 472.8 
        & 34.7350 & 549.9 \\ 

        DCTW-cache
        & 34.2132 & 135.5  
        & 33.8163 & 172.7
        & 34.1025 & 224.1 
        & 34.7246 & 140.6  
        & 34.3188 & 174.4 
        & 34.7350 & 226.2\\

        \bottomrule
    \end{tabular}
    \end{adjustbox}
\end{table}

\textbf{Results analysis.} The results in Table~\ref{tab:Q-rank} indicate that the effect of the $Q$ rank depends on the data type. Increasing $Q$ consistently improves the PSNR on the hyperspectral dataset \emph{mian}, but provides no consistent performance gain on the color video datasets \emph{container} and \emph{eagle}, where $Q=1$ generally already achieves good reconstruction accuracy. The three update strategies yield the same reconstruction quality but differ substantially in computational cost. Under the executable settings, DCTW with structural caching achieves speedups of approximately $41.8\text{--}201.8$ and $1.8\text{--}7.9$ over explicit construction and implicit contraction, respectively. Moreover, explicit construction runs out of memory under several high-rank settings on \emph{eagle}, whereas implicit contraction and structural caching remain executable. These results suggest that the $Q$ rank should be selected according to the characteristics of the data.

\section{Conclusion}\label{sect_6}
In this paper, we introduced the dual-core tensor wheel (DCTW)
decomposition, which replaces the high-order central core of TW with
two lower-order cores connected by a shared latent bond. This structure
retains the outer-ring interactions while providing a more flexible
representation of local and nonlocal correlations. Based on DCTW, we
developed PAM-based algorithms for tensor completion and robust tensor
denoising. The proposed structure-cached update further reuses
intermediate contractions to reduce redundant computation. Compared
with explicit and implicit updates, it reduces runtime and limits the
growth of additional peak memory as the ranks increase.

Experiments on color videos show that DCTW achieves the highest MPSNR across all completion tests and the highest MSSIM in most cases. In the denoising experiments, DCTW obtains the highest MPSNR and MSSIM at noise densities of $0.3$ and $0.4$ and remains competitive at a noise density of $0.5$. The structural comparison further shows that, under the same rank settings, DCTW achieves reconstruction performance comparable to or better than that of TW. Moreover, DCTW supports more flexible rank configurations, allowing its reconstruction performance to be further improved using higher rank settings. Overall, DCTW achieves a favorable balance among reconstruction accuracy, computational efficiency, and scalability. Future work will focus on automatic rank and structure selection, more memory-efficient tensor contractions, and improved robustness under extremely high noise levels.

%%%%%%%%%%%%%%%%%%%%%%%%%%%%%%%%%%%
% \section{Numerical experiments}\label{numerical} 

\section*{Acknowledgment}
\noindent{\bf Funding:} This research is supported by the National Natural Science Foundation of China (NSFC) grants 92473208, 12401415, the Key Program of National Natural Science of China 12331011,  the Natural Science Foundation of Hunan Province (No. 2025JJ60009), Project of Scientific Research Fund of Hunan Provincial Science and Technology Department (No. 25B0148), the Graduate Research Innovation Project of Xiangtan University (No. XDCX2026Y286).

\noindent{\bf Data Availability:} Enquiries about data/code availability should be directed to the authors.

\noindent{\bf Competing interests:} The authors have no competing interests to declare that are relevant to the content of this paper.

% \section{Conclusion}\label{conclusion}
   
%%%%%%%%%%%%%%%%%%%%%%%%

\appendix

%%%%%%%%%%%%%%%%%%%%%%%%%
\bibliographystyle{plain}	
\bibliography{refs}
%%%%%%%%%%%%%%%%%%%%%%%%%%%%	

\end{document}